\documentclass[10pt,twocolumn,letterpaper]{article}

\usepackage[pagenumbers]{cvpr} 

\usepackage{booktabs}
\usepackage{multirow}
\usepackage{graphicx}
\usepackage[normalem]{ulem}
\useunder{\uline}{\ul}{}
\newcommand{\myrule}{\specialrule{1pt}{.1pt}{.1pt}}
\newcommand{\mytinyrule}{\specialrule{.3pt}{.1pt}{.1pt}}

\usepackage[dvipsnames]{xcolor}

\newcommand{\trm} {\textrm}

\newcommand{\tbf} {\textbf}
\newcommand{\sbf} {\boldsymbol}
\newcommand{\mcal}{\mathcal}
\newcommand{\bn} {\trm{BN}}
\newcommand{\relu} {\trm{ReLU}}

\newcommand{\ssubsec}{\vspace{0pt}\noindent\tbf}

\definecolor{bg-nor}{HTML}{B4E5A2}
\definecolor{bg-abn}{HTML}{FDA9A9}

\definecolor{cvprblue}{rgb}{0.21,0.49,0.74}
\usepackage[pagebackref,breaklinks,colorlinks,citecolor=cvprblue]{hyperref}

\def\paperID{2757} 
\def\confName{CVPR}
\def\confYear{2024}

\title{BatchNorm-based Weakly Supervised Video Anomaly Detection}  

\author{
Yixuan Zhou\textsuperscript{1} \and Yi Qu\textsuperscript{1} \and Xing Xu\textsuperscript{1,}\thanks{Corresponding author.} \and Fumin Shen\textsuperscript{1} \and 
Jingkuan Song\textsuperscript{1} \and Hengtao Shen\textsuperscript{1,2} \\
\textsuperscript{1}Center for Future Media \& School of Computer Science and Engineering, \\
University of Electronic Science and Technology of China
\quad \textsuperscript{2}Peng Cheng Laboratory, China \\
{\tt\small yxzhou@std.uestc.edu.cn, iquyiiii@gmail.com, xing.xu@uestc.edu.cn, jingkuan.song@gmail.com} \\
{\tt\small fumin.shen@gmail.com, shenhengtao@hotmail.com}
}

\begin{document}
\maketitle
\begin{abstract}
    In weakly supervised video anomaly detection (WVAD), where only video-level labels indicating the presence or absence of abnormal events are available, the primary challenge arises from the inherent ambiguity in temporal annotations of abnormal occurrences.
    Inspired by the statistical insight that temporal features of abnormal events often exhibit outlier characteristics, we propose a novel method, BN-WVAD, which incorporates BatchNorm into WVAD.
    In the proposed BN-WVAD, we leverage the Divergence of Feature from Mean vector (DFM) of BatchNorm as a reliable abnormality criterion to discern potential abnormal snippets in abnormal videos.
    The proposed DFM criterion is also discriminative for anomaly recognition and more resilient to label noise, serving as the additional anomaly score to amend the prediction of the anomaly classifier that is susceptible to noisy labels.
    Moreover, a batch-level selection strategy is devised to filter more abnormal snippets in videos where more abnormal events occur.
    The proposed BN-WVAD model demonstrates state-of-the-art performance on UCF-Crime with an AUC of 87.24\%, and XD-Violence, where AP reaches up to 84.93\%.
    Our code implementation is accessible at \url{https://github.com/cool-xuan/BN-WVAD}.
\end{abstract}

\section{Introduction}

Video anomaly detection (VAD)~\cite{mehran2009abnormal,li2013anomaly} aims to detect and locate abnormal events in videos, which is of great importance in various real-world applications, such as intelligent surveillance~\cite{UCF} and autonomous driving~\cite{di2021pixel}. Yet, collecting a large-scale dataset with detailed temporal annotations of abnormal events is labor-intensive and time-consuming, which hinders the development of VAD. In recent years, \emph{weakly supervised video anomaly detection} (WVAD), requiring solely video-level labels denoting the presence or absence of abnormal events, has attracted increasing attention~\cite{RTFM,S3R,UR-DMU,CUNet,CLAWS,GCN,MSL,MIST,SAS} and outperformed unsupervised methods~\cite{GCL,FPDM} by a large margin.

In such a weakly supervised fashion, the principal challenge of WVAD stems from the lack of temporal annotations for abnormal events.
To address this challenge, existing methods resort to certain abnormality criteria, such as feature magnitude~\cite{S3R,RTFM}, or attention~\cite{UR-DMU}, to identify top-$k$ potential abnormal snippets in labeled abnormal videos.
These selected snippets also serve as pseudo temporal annotations~\cite{CUNet,GCN}, which are expected to provide supervision for distinguishing abnormal events from normal ones.
Rather than directly training on pseudo temporal annotations, existing methods~\cite{RTFM,S3R,UR-DMU} draw inspiration from Multi-Instance Learning (MIL)~\cite{MIL,li2015multiple} to improve the tolerance to the presence of mislabeled snippets.
Specifically, the selected abnormal snippets are gathered as the positive bag and paired with a negative bag constructed from normal videos to train the anomaly classifier.

\begin{figure}[t]
    \begin{subfigure}{0.495\linewidth}
        \centering
        \includegraphics[width=\linewidth]{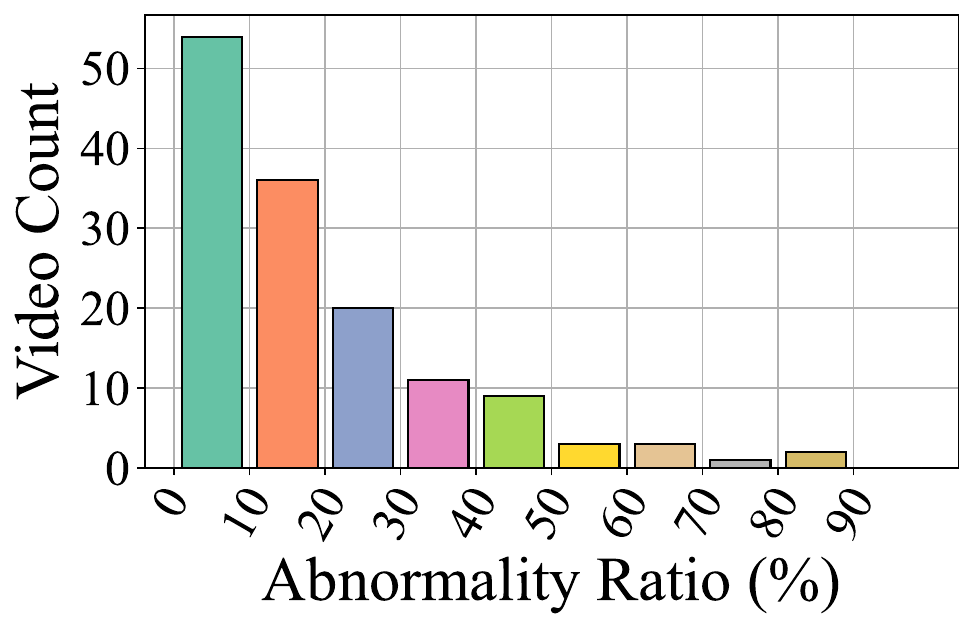}
        \caption{UCF-Crime~\cite{UCF}}
        \label{fig:anomalyRatio-ucf}
    \end{subfigure}
    \begin{subfigure}{0.495\linewidth}
        \centering
        \includegraphics[width=\linewidth]{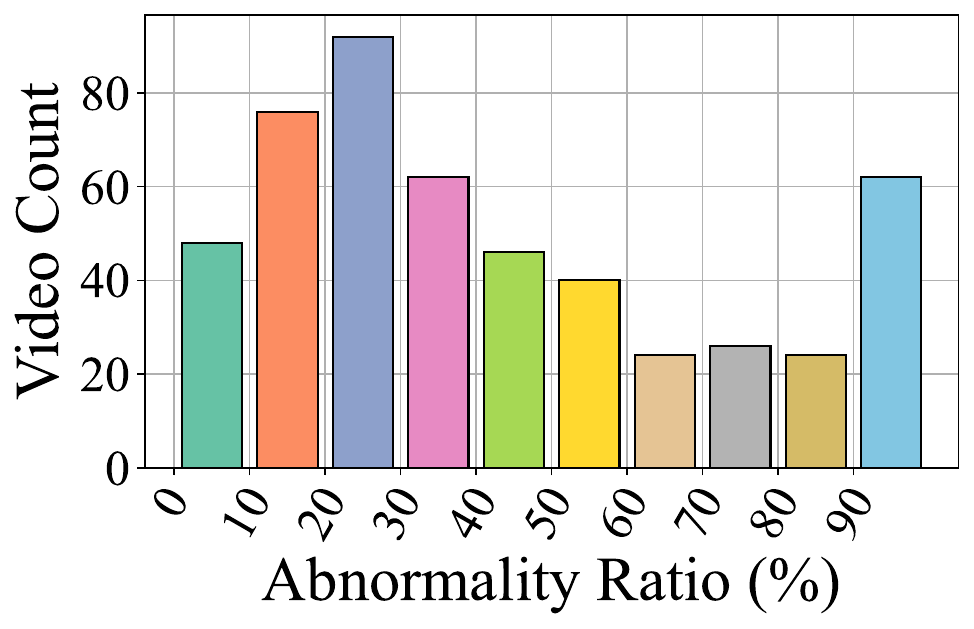}
        \caption{XD-Violence~\cite{XD}}
        \label{fig:anomalyRatio-xd}
    \end{subfigure}
    \caption{The illustration of the abnormality ratio distribution of test sets. The abnormality ratio varies across different videos, especially in XD-Violence~\cite{XD} with higher abnormality ratios.}
    \label{fig:anomaly_ratio}
    \vspace{-3pt}
\end{figure}

Although existing methods~\cite{RTFM,S3R,UR-DMU,CUNet} have demonstrated promising performance, they still suffer from three primary limitations.
1) \emph{Unreliable abnormality criteria.}
Previous abnormality criteria primarily rely on some assumptions~\cite{RTFM} or black-box models~\cite{UR-DMU,CUNet}, leading to less reliable pseudo temporal annotations. 
For instance, the widely employed feature magnitude~\cite{RTFM, S3R} is based on a plausible assumption that abnormal snippets exhibit a larger feature magnitude than normal snippets. However, the mere reliance on large feature magnitudes does not guarantee sufficient discrimination for abnormal snippets.
2) \emph{Limitation of sample-level selection strategy.}
Previous methods~\cite{RTFM,UR-DMU,S3R} select top-$k$ potential abnormal snippets for each video, without considering the varying abnormality ratio across different videos as shown in Fig.~\ref{fig:anomaly_ratio}, where the abnormality ratio is defined as the proportion of abnormal snippets in each video.
Uniformly selecting potential abnormal snippets in each video may neglect significant abnormal snippets in videos with higher abnormality ratios, thus missing instructive supervision for anomaly recognition.
3) \emph{Sensitivity to the misselection in abnormal video.}
Misselection of abnormal snippets is inevitable in WVAD, introducing label noise in pseudo temporal annotations. Despite the adoption of MIL, the anomaly classifier remains susceptible to label noise, trapped in the dilemma of recognizing mislabeled `abnormal' snippets.

\begin{figure}
    \centering
    \includegraphics[width=\linewidth]{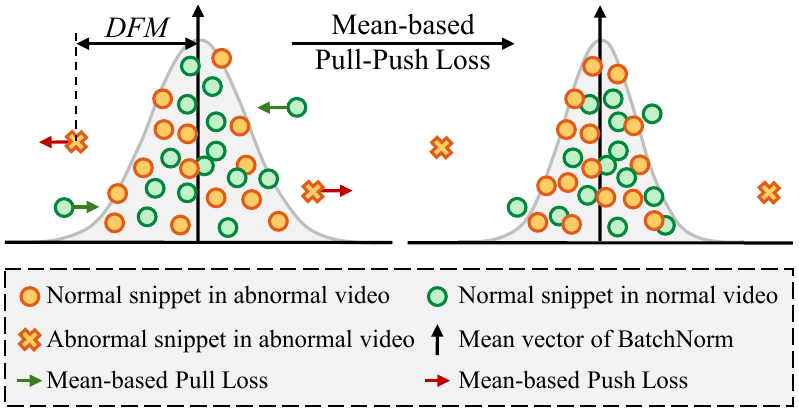}
    \caption{Intuition of the proposed DFM criterion and Mean-based Pull-Push (MPP) loss. The mean vector of BatchNorm is regarded as a statistical reference to separate potential abnormal and normal snippets, and MPP Loss encourages their separation.}
    \label{fig:intro}
    \vspace{-3pt}
\end{figure}

By introducing the statistical principles underlying BatchNorm~\cite{BatchNorm}, we propose a novel BatchNorm-based WVAD model, dubbed BN-WVAD, to tackle the above limitations.
From the statistical perspective, we observe that temporal features of abnormal events often exhibit characteristics of outliers~\cite{wilson2023safe,sun2021react} concerning the mean vector computed by BatchNorm, which predominantly captures the normality of the feature distribution~\cite{AnoOnly}.
In other words, the mean vector of BatchNorm can be regarded as a reference to distinguish potential abnormal snippets from normal ones.
Accordingly, our BN-WVAD introduces the Divergence of Feature from Mean (DFM) as a novel abnormality criterion to supersede existing ones~\cite{RTFM,S3R,UR-DMU,CUNet}, discerning reliable potential abnormal snippets.
Furthermore, we propose a Mean-based Pull-Push (MPP) loss to enhance the separation of DFM for abnormal features compared to normal features, as illustrated in Fig.~\ref{fig:intro}.

\begin{figure*}[t]
    \centering
    \includegraphics[width=0.97\textwidth]{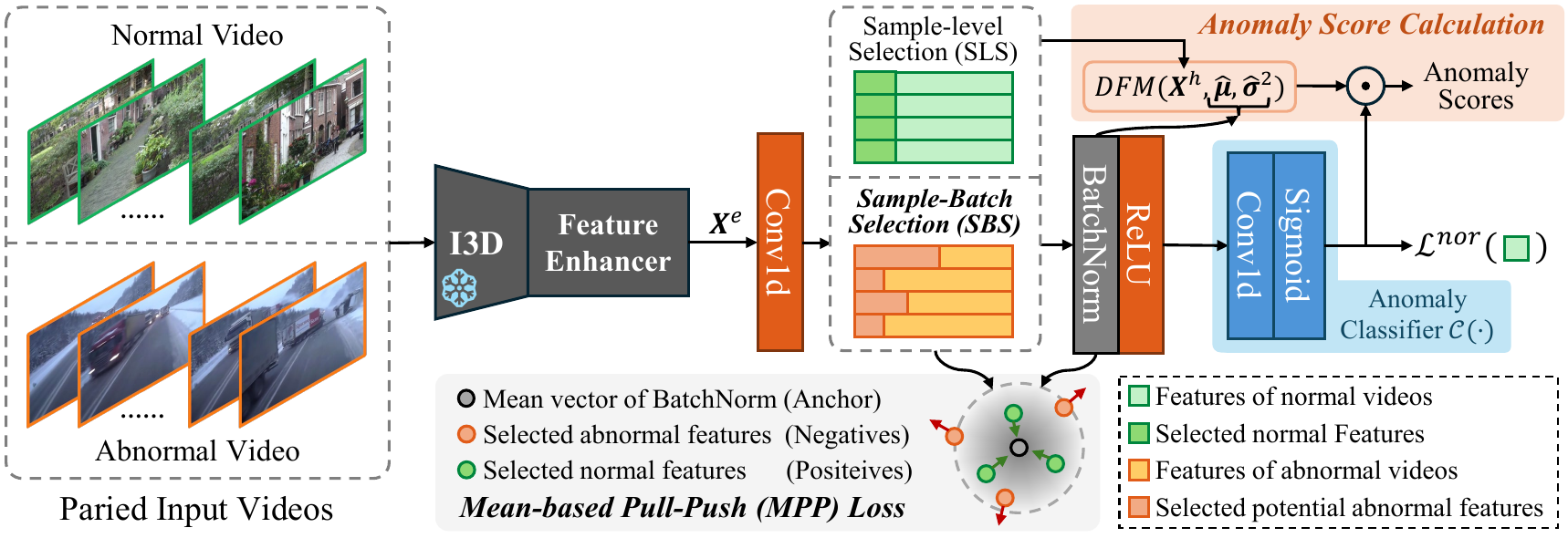}
    \caption{The overall framework of our proposed BN-WVAD model. The input mini-batch is composed of half normal videos and half abnormal videos and embedded by a frozen I3D~\cite{I3D} followed by a Transformer-based enhancer~\cite{UR-DMU}, yielding enhanced features $\sbf{X}^e$. In particular, the visualized features are sorted by the DFM criterion (Eq.~\ref{eq:dfm}) in descending order, for the convenience of illustrating the different selection strategies adopted in normal and abnormal videos. Only one hidden feature $\sbf{X}^{h}$ is visualized here for better illustration.
    }
    \label{fig:method-overall}
    \vspace{-3pt}
\end{figure*}

To overcome the limitation of sample-level top-$k$ selection, we draw inspiration from the focus of BatchNorm and introduce a Batch-level Selection (BLS) strategy to filter more potential abnormal snippets in the video with a higher occurrence of abnormal events.
A Sample-Batch Selection (SBS) strategy is further devised to combine the advantages of sample-level and batch-level selection strategies.
To enhance the tolerance to mislabeled abnormal snippets, we only train the vulnerable anomaly classifier in our BN-WVAD on certainly normal snippets from normal videos, mitigating confusion induced by label noise.
Additionally, the proposed DFM criterion serves as the other discrimination criterion, which is acquired in the dense feature space and proves more resilient to the misselection~\cite{wu2020topological,qu2021dat}. As shown in Fig.~\ref{fig:method-overall}, the final anomaly scores in the proposed BN-WVAD are calculated by aggregating the DFM scores with the prediction of the anomaly classifier.

Our BN-WVAD is a straightforward yet effective model, surpassing existing methods~\cite{RTFM,S3R,UR-DMU,CUNet,GCN,MSL,CLAWS,MIST,SAS} and achieving SOTA performance on UCF-Crime~\cite{UCF} and XD-Violence~\cite{XD}.
Notably, the flexibility of incorporating our DFM criterion and BLS strategy~\cite{RTFM,UR-DMU} 
The insight that dense feature space exhibits increased robustness to misselection is also instructive for future research.
Our main contributions are summarized as follows:
\begin{itemize}
    \item We introduce a novel BatchNorm-based WVAD model termed BN-WVAD, where our DFM criterion plays a crucial role in screening reliable abnormal snippets. 
    The MPP loss is further proposed to gather normal features and enlarge DFM of potential abnormal features.
    \item Inspired by the introduction of BatchNorm, we devise a sample-batch selection strategy to fully exploit instructive abnormal snippets within abnormal videos.
    \item Our BN-WVAD calculates the final anomaly scores by aggregating the DFM scores with the prediction of the anomaly classifier, where the proposed DFM criterion is discriminative and more resilient to label noise.
\end{itemize}

\section{Related Work}

\textbf{Unsupervised video anomaly detection.}
Restricted to the difficulty of collecting and annotating large-scale abnormal videos~\cite{UCF}, unsupervised video anomaly detection (UVAD)~\cite{li2013anomaly} has been widely studied in the early years.
Due to the only availability of normal videos, UVAD methods mainly focus on learning the normality and detecting abnormal events by identifying the deviations from normality, which is also deemed as the one-class classification problem~\cite{ruff2018deep}.
The representative methods can be roughly divided into two categories: reconstruction-based methods~\cite{GCL,FPDM,sabokrou2018adversarially,nguyen2019anomaly,park2020learning,liu2021hybrid,ShanghaiTec} and regression-based methods~\cite{pang2020self,park2020learning,abati2019latent,hirschorn2023normalizing,georgescu2021anomaly,markovitz2020graph}.
The former focuses on learning normal video representations by reconstruction, and the latter uses self-training~\cite{pang2020self, georgescu2021anomaly} to grasp the normality.
However, the performance of UVAD methods is suboptimal due to the absence of abnormal videos during training.

\noindent\textbf{Weakly supervised video anomaly detection.}
Although the fine-grained temporal annotations are impractical to obtain, the video-level labels are relatively feasible to annotate~\cite{UCF}.
With the practical accessibility of video-level labels~\cite{UCF, XD}, weakly supervised video anomaly detection (WVAD) has gained increasing attention in recent years.
Relying solely on video-level labels, existing WVAD methods~\cite{RTFM, S3R, UCF, UR-DMU, MSL, MIST, XD} often employ Multi-Instance Learning (MIL)~\cite{MIL,li2015multiple}. They train anomaly classifiers using positive (abnormal) and negative (normal) bags, generated based on tailored abnormality criteria such as feature magnitude~\cite{RTFM}.
Despite the absence of MIL in other methods~\cite{GCN,CLAWS,CUNet}, they still heavily depend on abnormality criteria for pseudo temporal annotation generation.
Although these methods have achieved promising results, they grapple with the unreliability of abnormality criteria and the limitation of top-$k$ selection strategy.
Introducing the statistical notion of BatchNorm, we propose the novel DFM criterion to measure the abnormality of snippets and a batch-level selection strategy to address the shortcoming of top-$k$ selection strategy in overlooking abnormal snippets from videos with high abnormality ratios.

\noindent\textbf{Normality modeling.}
In WVAD, since the ambiguity of temporal labels of abnormal events, normality modeling of definitely normal features in normal videos is of great importance.
Besides most methods~\cite{RTFM,GCL} embedding the knowledge of normality into the anomaly classifier, some methods~\cite{park2020learning,liu2021hybrid,UR-DMU} explicitly model the prototypes of normality into the additional memory module. Although there is no explicit normality modeling in our method, BatchNorm~\cite{BatchNorm} serves as a simple memory module to store the normality of the feature distribution.
Specifically, the mean vector computed by BatchNorm is statistically proved to be a good representation of normality~\cite{AnoOnly}, due to the overwhelming majority of normal snippets in videos. In particular, BatchNorm also spontaneously gathers the normal features in abnormal videos, which are neglected by previous methods~\cite{UR-DMU}. Therefore, the mean vector of BatchNorm can be regarded as a statistical reference to separate potential abnormal and normal snippets.

\section{The Proposed BN-WVAD Model}

In weakly supervised anomaly detection (WVAD), a training set consists of $N$ untrimmed videos $\mcal{V}=\{V_i\}_{i=1}^N$, where each video $V_i$ is associated with a video-level label $\mcal{Y}_i \in \{0,1\}$ denoting the absence or presence of abnormal events.
Correspondingly, the training set can be divided into two subsets: a normal set $\mcal{V}^n$=$\{V^n_i\}_{i=1}^{N^n}$ and an abnormal set $\mcal{V}^a$=$\{V^a_i\}_{i=1}^{N^a}$, where $N^n+N^a$=$N$.
In practice, the raw videos $\{V_i\}_{i=1}^N$ are beforehand encoded to snippet features $\{\sbf{X}_i\}_{i=1}^N$ using pre-trained backbones~\cite{I3D,C3D,VSwin,TSN,ResNeXt}.
A feature enhancer~\cite{UR-DMU} is applied to enhance the feature representation, resulting in $\sbf{X}^e$.
Our method operates based on these enhanced features, and the overall framework of our BN-WVAD model is depicted in Fig.~\ref{fig:method-overall}.

We first address the underestimated significance of BatchNorm in WVAD in Sec.~\ref{sec:method-bn}, which motivates us to propose our novel BatchNorm-based WVAD model.
Subsequently, we elaborate on the key components of our BN-WVAD model, including the DFM criterion in Sec.~\ref{sec:method-dfm}, the SBS strategy in Sec.~\ref{sec:method-selection}, and our specific anomaly score calculation in Sec.~\ref{sec:method-score}. Finally, we present the overall training objective in Sec.~\ref{sec:method-objective}.

\subsection{The Significance of BatchNorm in WVAD} \label{sec:method-bn}
Besides the well-known effect of BatchNorm~\cite{BatchNorm} in improving training stability and model generalization, its inherent superiority in statistical modeling of normality is underestimated in WVAD.
Consider a mini-batch of $B$ videos, the hidden features $\sbf{X}^h \in \mathbb{R}^{B \times T \times C}$ are fed into the BatchNorm layer, where $C$ denotes the dimension of $\sbf{X}^h$.
During training, the mean vector $\sbf{\mu} \in \mathbb{R}^{C}$ is automatically computed by BatchNorm as follows:
\begin{equation}
    \sbf{\mu} = \mathbb{E}(\sbf{X}^{h}) = \frac{1}{B \times T} \sum_{b=1}^B \sum_{t=1}^T \sbf{X}^{h}[b,t],
\end{equation}
where $\mathbb{E}(\cdot)$ denotes the expectation operator and $\sbf{X}^{h}[b,t]$ denotes the $t$-th snippet of the $b$-th video in the mini-batch.
In typical WVAD implementations~\cite{RTFM,S3R,UR-DMU,CUNet}, each mini-batch is constructed with an equal distribution of normal and abnormal videos, ensuring the majority of normal snippets.
Therefore, the mean vector $\sbf{\mu}$ is primarily determined by sufficient normal representations, in other words, capturing the normality of the feature distribution~\cite{AnoOnly}.
Importantly, BatchNorm naturally aggregates normal features even from abnormal videos, which are neglected by previous explicit memory modules~\cite{UR-DMU,liu2021hybrid}.

Furthermore, $\sbf{\mu}$ is derived from $B\times T$ snippet features, leading to the distribution in mini-batch that is statistically proven to follow a normal distribution, as asserted by the Central Limit Theorem (CLT)~\cite{CLT}. 
From this perspective, the features of abnormal snippets are more likely to be outliers and exhibit a notable divergence from the mean vector.
This observation motivates the introduction of the DFM criterion, which plays a central role in our BN-WVAD model.

\subsection{BatchNorm-based Abnormality Criterion} \label{sec:method-dfm}
Following our insight into statistical normality modeling of BatchNorm in WVAD, the mean vector $\sbf{\mu}$ can be regarded as a statistical representation of normality to distinguish potential abnormal snippets from normal ones.
As a result, our BN-WVAD utilizes the Divergence of Feature from Mean (DFM) vector of BatchNorm as a novel abnormality criterion to substitute the plausible ones~\cite{RTFM,S3R,UR-DMU,CUNet}, improving the reliability of abnormal snippet selection.

Accommodating the anisotropic Gaussian distribution among different dimensions of the multivariate feature space~\cite{ghorbani2019mahalanobis}, we employ the Mahalanobis distance~\cite{de2000mahalanobis} to quantify the divergence between the hidden features $\sbf{X}^h$ and the mean vector $\sbf{\mu}$ computed by BatchNorm.
Specifically, the proposed DFM is formulated as follows:
\begin{align} \label{eq:dfm}
    \trm{DFM} & (\sbf{X}^{h}[b,t], \sbf{\mu}, \sbf{\sigma}^2) \nonumber                                               \\
              & = \sqrt{(\sbf{X}^{h}[b,t] - \sbf{\mu})^{\trm{T}} {\sbf{\Sigma}}^{-1} (\sbf{X}^{h}[b,t] - \sbf{\mu})},
\end{align}
where $\sbf{\Sigma} \in \mathbb{R}^{C \times C}$ is the covariance matrix of the hidden features $\sbf{X}^h$, presented as $\sbf{\Sigma} = \trm{diag}(\sbf{\sigma}^2)$ with $\sbf{\sigma}^2 \in \mathbb{R}^{C}$ being the variances of each dimension.

Notably, in common practice~\cite{paszke2017automatic, abadi2016tensorflow}, the running mean vector $\sbf{\hat{\mu}}$ and variance vector $\sbf{\hat{\sigma}}^2$ are updated according to exponential moving average (EMA)~\cite{gardner1985exponential} with momentum $\alpha=0.1$ as follows:
\begin{align}
    \sbf{\hat{\mu}}      & = (1-\alpha) \sbf{\hat{\mu}} + \alpha \sbf{\mu},           \\
    \sbf{\hat{\sigma}}^2 & = (1-\alpha) \sbf{\hat{\sigma}}^2 + \alpha \sbf{\sigma}^2.
\end{align}
The EMA-based statistics $\sbf{\hat{\mu}}$ and $\sbf{\hat{\sigma}}^2$ capture the long-term statistics of the feature distribution, avoiding potential bias that may arise from statistics calculated in mini-batches.
Compared with the statistics derived from mini-batches, running statistics are more representative of normality and more robust to the presence of abnormal features.
Furthermore, the utilization of $\sbf{\hat{\mu}}$ and $\sbf{\hat{\sigma}}^2$ maintains the consistency between training and testing.

To encourage the divergence between the potential abnormal features with the mean vector $\sbf{\hat{\mu}}$, and to gather the normal features, we propose an incident Mean-based Pull-Push (MPP) loss for optimization.
In particular, according to our DFM criterion, $K$ potential abnormal features in abnormal videos and $K$ normal features in normal videos with the largest $K$ DFM scores are selected from $\sbf{X}^{h}$ and denoted as $\sbf{X}^{a}_\trm{dfm} \in \mathbb{R}^{K \times C}$ and $\sbf{X}^{n}_\trm{dfm} \in \mathbb{R}^{K \times C}$, respectively.
Borrowing the intuition of Triplet loss~\cite{TripletLoss}, we treat the mean vector $\sbf{\hat{\mu}}$ as the only anchor, selected normal features $\sbf{X}^{n}_\trm{dfm}$ as positives, and selected abnormal features $\sbf{X}^{a}_\trm{dfm}$ as negatives.
Correspondingly, based on the DFM criterion (Eq.~\ref{eq:dfm}),
the proposed MPP loss is formulated as follows:
\begin{align} \label{eq:mpp}
    \mcal{L}^{\trm{mpp}}(\sbf{X}^{n}_\trm{dfm}, \sbf{X}^{a}_\trm{dfm},\sbf{\hat{\mu}}, & \sbf{\hat{\sigma}}^2) \nonumber                                                         \\
    = \frac{1}{K} \sum_{k=1}^K [m                                                      & + \trm{DFM}(\sbf{X}^n_\trm{dfm}[k], \sbf{\hat{\mu}}, \sbf{\hat{\sigma}}^2)              \\
                                                                                       & - \trm{DFM}(\sbf{X}^a_\trm{dfm}[k], \sbf{\hat{\mu}}, \sbf{\hat{\sigma}}^2)],  \nonumber
\end{align}
where $m$ is the margin, set to 1 in our BN-WAVD implementation, to enlarge the separation of $\sbf{X}^{n}_\trm{dfm}$ and $\sbf{X}^{a}_\trm{dfm}$.

\subsection{Sample-Batch Selection (SBS) Strategy} \label{sec:method-selection}

In addition to the prevalent sample-level selection (SLS) strategy~\cite{RTFM,S3R,UR-DMU,CUNet} in WVAD, our BN-WVAD also incorporates the statistical notion of BatchNorm into abnormal snippet selection, introducing a batch-level selection (BLS) strategy.
Drawing inspiration from our insight into the statistical modeling of BatchNorm, we conjecture that, despite the varying abnormality ratio across different videos, the overall abnormality ratio of the entire mini-batch is relatively stable.
Hence, the proposed BLS strategy screens the potential abnormal snippets within each mini-batch rather than each video, which is more flexible to the unequal abnormality ratio distribution.

\begin{figure}
    \centering
    \begin{subfigure}{0.32\linewidth}
        \centering
        \includegraphics[width=\linewidth]{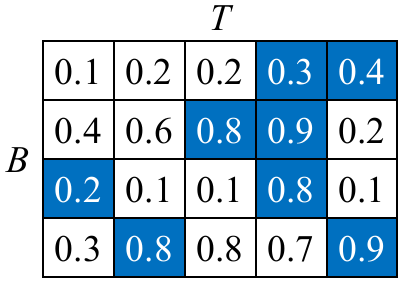}
        \caption{Sample-level Selection}
        \label{fig:selection-sls}
    \end{subfigure}
    \begin{subfigure}{0.32\linewidth}
        \centering
        \includegraphics[width=\linewidth]{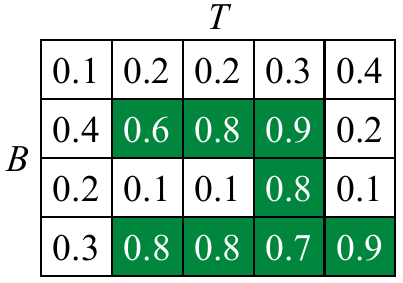}
        \caption{Batch-level Selection}
        \label{fig:selection-bls}
    \end{subfigure}
    \begin{subfigure}{0.34\linewidth}
        \centering
        \includegraphics[width=0.95\linewidth]{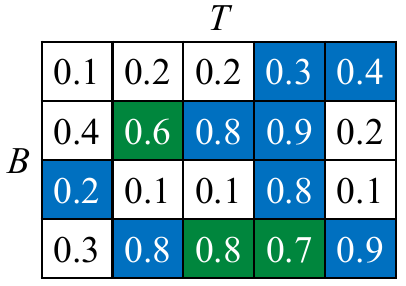}
        \caption{Sample-Batch Selection}
        \label{fig:selection-sbs}
    \end{subfigure}
    \caption{Illustration of different selection strategies adopted in $B$=4 abnormal videos with $T$=5 snippets. Values in boxes denote the abnormality criterion score of each snippet. Both sample-level selection ratio $\rho_s$ and batch-level selection ratio $\rho_b$ are set to 40\%.}
    \label{fig:method-selection}
    \vspace{-3pt}
\end{figure}

Specifically, two selection ratios, denoted as $\rho_s$ and $\rho_b$, are introduced to regulate the proportion of selected abnormal snippets within each video and mini-batch, respectively. For intuitive illustration, we assume that the mini-batch is composed of $B$=4 abnormal videos with $T$=5 snippets, and both $\rho_s$ and $\rho_b$ are set to 40\% in Fig.~\ref{fig:method-selection}.
When only the SLS strategy is adopted, the abnormal snippets with large abnormality scores (0.8 and 0.7) in the 4th video are ignored, as illustrated in Fig.~\ref{fig:selection-sls}.
In contrast, the proposed BLS strategy filters abnormal snippets from the perspective of statistics and successfully captures all 4 potential abnormal snippets in the 4th video.
However, when facing the inconspicuous abnormal snippets with relatively small abnormality scores (0.3 and 0.4) in the 1st video, our BLS strategy fails to discern them.

We further devise a Sample-Batch Selection (SBS) strategy to complement the disadvantages of SLS in insufficient selection and BLS being insensitive to hard abnormal snippets.
As depicted in Fig.~\ref{fig:selection-sbs}, our SBS strategy considers the union of selected snippets from SLS and BLS as the final selection.
Particularly, our BN-WVAD model adopts the introduced SBS strategy in abnormal videos, while exclusively applying SLS strategy to normal videos.
The overall selection amount in normal videos is equal to the number of selected abnormal snippets according to SBS strategy, enabling the computation of our pairwise MPP loss (Eq.~\ref{eq:mpp}).

\subsection{Anomaly Score Calculation} \label{sec:method-score}
In existing methods~\cite{RTFM,S3R,UR-DMU,CUNet}, the ultimate anomaly scores are directly derived from the predictions of the anomaly classifier $\mcal{C}(\cdot)$. 
However, the anomaly classifier remains susceptible to the misselection of abnormal snippets, even with the adoption of MIL, resulting in potential misclassifications of normal snippets as abnormal instances.
In our BN-WAVD, the anomaly classifier is solely trained on the certainly normal snippets from normal videos, eliminating the confusion induced by label noise. 
Instead of using bag-level binary cross-entropy loss, we employ a snippet-level regression loss to supervise the anomaly classifier $\mcal{C}(\cdot)$ with hidden features $\sbf{X}^n \in \mathbb{R}^{\frac{B}{2} \times T \times C}$ of normal videos, formulated as follows:
\begin{equation}
    \mcal{L}^{\trm{nor}}(\sbf{X}^n; \mcal{C}) = \sum_{b=1}^{B/2} \| \mcal{C}(\relu(\bn(\sbf{X}^n[b]))) \Vert_2,
\end{equation}
where $\| \cdot \Vert_2$ denotes the $L_2$ norm, and $\sbf{X}^n[b]$ denotes the hidden feature of the $b$-th video in the normal mini-batch.
Notably, predicted scores of all $T$ snippets are used for supervision, without the need for hard sample mining like previous methods~\cite{RTFM,S3R,UR-DMU}.

To enhance the discriminative capacity further, we introduce our DFM criterion (Eq.~\ref{eq:dfm}) as an additional discrimination criterion, which is acquired in dense feature space and exhibits increased robustness to label noise~\cite{qu2021dat}.
The final anomaly scores are calculated by combining DFM scores and the prediction of the anomaly classifier as follows:
\begin{equation} \label{eq:score}
    \trm{Score} = \mcal{C}(\relu(\bn(\sbf{X}^h))) * \trm{DFM}(\sbf{X}^h, \sbf{\hat{\mu}}, \sbf{\hat{\sigma}}^2),
\end{equation}
where the DFM scores and the prediction of the anomaly classifier are aggregated by element-wise multiplication `$*$'.

\subsection{The Training Objective} \label{sec:method-objective}
As for implementation, we utilize two Conv1d layers to obtain the hidden features $\sbf{X}^{h1}$ and $\sbf{X}^{h2}$, followed by BatchNorm and ReLU. Both $\sbf{X}^{h1}$ and $\sbf{X}^{h2}$ are supervised by the proposed MPP loss (Eq.~\ref{eq:mpp}), and their DFM criterion (Eq.~\ref{eq:dfm}) values are summed up for the final anomaly score calculation (Eq.~\ref{eq:score}).
The overall loss objective of our method is formulated as follows:
\begin{equation}
    \mcal{L} = \mcal{L}^{\trm{nor}} + \lambda_1 \mcal{L}^{\trm{mpp}}_1 + \lambda_2 \mcal{L}^{\trm{mpp}}_2,
\end{equation}
where $\mcal{L}^{\trm{mpp}}_1$ and $\mcal{L}^{\trm{mpp}}_2$ denote the MPP losses calculated on $\sbf{X}^{h1}$ and $\sbf{X}^{h2}$, respectively. $\lambda_1$ and $\lambda_2$ are the hyper-parameters to balance the loss terms.

\section{Experiments}

\subsection{Datasets and Evaluate Protocols}

\tbf{Datasets.}  We evaluate our proposed BN-WVAD on two prominent WVAD datasets: UCF-Crime~\cite{UCF} and XD-Violence~\cite{XD}, where video-level labels are accessible.

\emph{UCF-Crime} collects 1900 real-world surveillance videos annotating 13 types of anomalous events, e.g., abuse, robbery, explosion, and road accidents. In the training set with video-level labels only, there are 800 normal and 810 abnormal videos. The testing set comprises 140 normal and 150 abnormal videos with temporal annotations for the evaluation of frame-level anomaly detection.

\emph{XD-Violence} is a multisource dataset, collected from movies, surveillance cameras, etc. It is the largest WVAD dataset with video-level labels available, composed of 4754 untrimmed videos with 6 types of anomalous events, e.g., abuse, car accidents, and shootings. The training set includes 2049 normal and 1905 abnormal videos, labeled at the video level. The testing set, with frame-level labels, consists of 300 normal and 500 abnormal videos. Notably, both video and audio data are available in XD-Violence.

\ssubsec{Evaluation protocols.}
We adhere to established evaluation protocols to ensure fair comparisons with previous methods. Specifically, we utilize the area under the curve (AUC) of the frame-level receiver operating characteristic (ROC) curve as the primary metric for UCF-Crime. On XD-Violence, frame-level average precision (AP) is the key metric for assessment.

\begin{table}[]
    \centering
    \resizebox{0.95\columnwidth}{!}{%
        \begin{tabular}{c|lccc}
            \myrule
            \multicolumn{1}{l|}{}                           & Method                                 & Venue     & Feature        & AUC (\%) \\ \mytinyrule
            \multirow{2}{*}{\rotatebox{270}{Un.\quad}}      & GCL~\cite{GCL}                         & CVPR$'$23 & ResNeXt        & 74.20    \\
                                                            & FPDM~\cite{FPDM}                       & ICCV$'$23 & Image          & 74.70    \\ \mytinyrule
            \multirow{13}{*}{\rotatebox{270}{Weakly\qquad}} & Sultani et al.~\cite{UCF}              & CVPR$'$18 & C3D            & 75.41    \\
                                                            & Sultani et al.~\cite{UCF}              & CVPR$'$18 & I3D            & 76.21    \\
                                                            & GCN~\cite{GCN}                         & CVPR$'$19 & TSN            & 82.12    \\
                                                            & HL-Net~\cite{XD}                       & ECCV$'$20 & I3D            & 82.44    \\
                                                            & CLAWS~\cite{CLAWS}                     & ECCV$'$20 & C3D            & 83.03    \\
                                                            & MIST~\cite{MIST}                       & CVPR$'$21 & I3D            & 82.30    \\
                                                            & RTFM~\cite{RTFM}                       & ICCV$'$21 & I3D            & 84.30    \\
                                                            & MSL~\cite{MSL}                         & AAAI$'$22 & I3D            & 85.30    \\
                                                            & S3R~\cite{S3R}                         & ECCV$'$22 & I3D            & 85.99    \\
                                                            & SAS~\cite{SAS}                         & arXiv$'$23 & I3D           & 86.19    \\
                                                            & CU-Net~\cite{CUNet}                    & CVPR$'$23 & I3D            & 86.22    \\
                                                            & UR-DMU~\cite{UR-DMU}                   & AAAI$'$23 & I3D            & 86.97    \\
                                                            & UR-DMU$^\dagger$\cite{UR-DMU}          & AAAI$'$23 & I3D            & 86.23    \\ \cline{2-5}
                                                            & \multicolumn{2}{c}{BN-WVAD (\textbf{Ours})} & I3D       & \textbf{87.24}            \\ \myrule
        \end{tabular}%
    }
    \caption{Comparison of AUC (\%) on UCF-Crime~\cite{UCF}. The methods are divided into two categories: unsupervised (Un.) and weakly supervised (Weakly). `$\dagger$' denotes the reproduced results of open-source code~\cite{UR-DMU} by ourselves.}
    \label{tab:comparison-ucf}
    \vspace{-3pt}
\end{table}

\subsection{Implementation Details}
Following the previous SOTA UR-DMU~\cite{UR-DMU}, our BN-WVAD employs the I3D~\cite{I3D} to extract the snippet features. For the XD-Violence, raw audio is embedded as audio features through VGGish~\cite{VGGish}. The untrimmed video features are linearly interpolated to a standardized length of 200 snippets. We leverage a Transformer-based enhancer~\cite{UR-DMU} to enhance feature representation, with an output dimension of 512.
Upon the enhanced features, our BN-WVAD utilizes two Conv1d layers to obtain the hidden features $\sbf{X}^{h1}$ and $\sbf{X}^{h2}$, followed by BatchNorm and ReLU.
The kernel size of two Conv1d layers is set to 1, and the output dimension is 32 and 16. The hyper-parameters $\lambda_1$ and $\lambda_2$ are set to 5 and 20, respectively. Two selection ratios are set according to the abnormality distribution of datasets, i.e., $\rho_s=0.1$ and $\rho_b=0.2$ on UCF-Crime, and $\rho_s=0.2$ and $\rho_b=0.4$ on XD-Violence.
We utilize Adam~\cite{kingma2014adam} as the optimizer with a learning rate of 0.0001 and a weight decay of 0.00005. The model is trained for 3000 iterations with the mini-batch of 64 normal and abnormal videos. During inference, the multi-crop aggregation is adopted to obtain the final anomaly scores, where the number of crops is set to 10 for UCF-Crime and 5 for XD-Violence.

\begin{table}[]
    \centering
    \resizebox{0.92\columnwidth}{!}{%
        \begin{tabular}{lccc}
            \myrule
            Method                                 & Venue      & Feature        & AP (\%) \\ \mytinyrule
            Sultani et al.~\cite{UCF}              & CVPR$'$18  & I3D            & 73.20   \\
            HL-Net~\cite{XD}                       & ECCV$'$20  & I3D            & 73.67   \\
            HL-Net~\cite{XD}                       & ECCV$'$20  & I3D+VGGish     & 78.64   \\
            RTFM~\cite{RTFM}                       & ICCV$'$21  & I3D            & 77.81   \\
            MSL~\cite{MSL}                         & AAAI$'$22  & I3D            & 78.28   \\
            S3R~\cite{S3R}                         & ECCV$'$22  & I3D            & 80.26   \\
            CU-Net~\cite{CUNet}                    & CVPR$'$23  & I3D            & 78.74   \\
            CU-Net~\cite{CUNet}                    & CVPR$'$23  & I3D+VGGish     & 81.43   \\
            UR-DMU~\cite{UR-DMU}                   & AAAI$'$23  & I3D            & 81.66   \\
            UR-DMU~\cite{UR-DMU}                   & AAAI$'$23  & I3D+VGGish     & 81.77   \\
            MACIL-SD~\cite{MACIL-SD}               & MM$'$22    & I3D+VGGish     & 83.40   \\
            SAS~\cite{SAS}                         & arXiv$'$23 & I3D            & 83.59   \\ \mytinyrule
            \multicolumn{2}{c}{BN-WVAD (\textbf{Ours})} & I3D        & \textbf{84.93}           \\
            \multicolumn{2}{c}{BN-WVAD (\textbf{Ours})} & I3D+VGGish & \textbf{85.26}           \\ \myrule
        \end{tabular}%
    }
    \caption{Comparison of AP (\%) on XD-Violence~\cite{XD}. `+VGGish' refers to the methods with audio features as additional inputs.}
    \label{tab:comparison-xd}
    \vspace{-3pt}
\end{table}

\subsection{Comparison with SOTA Methods}

\tbf{UCF-Crime.} On this real-world surveillance dataset~\cite{UCF}, we compare our BN-WVAD with previous SOTA methods under unsupervised~\cite{GCL,FPDM} and weakly supervised~\cite{UCF,GCN,XD,CLAWS,MIST,RTFM,S3R,CUNet,UR-DMU} fashions, as reported in Table~\ref{tab:comparison-ucf}.
Leveraging video-level labels in WVAD proves advantageous, leading to a significant performance gap compared to UAD methods.
Compared with previous weakly supervised methods, the proposed BN-WVAD further improves the AUC score to 87.24\%.
Despite achieving a modest improvement of 0.27\% compared to the reported results of UR-DMU~\cite{UR-DMU}, our performance gain is commendable, especially when contrasted with the result (86.23\%) of our reproduction based on the official code \footnote{https://github.com/henrryzh1/UR-DMU}.

\ssubsec{XD-Violence.} Table~\ref{tab:comparison-xd} showcases the AP scores of video-only methods~\cite{RTFM,MSL,S3R,CUNet,UR-DMU,SAS} and audio-visual methods~\cite{XD,MACIL-SD} on this multi-modal dataset~\cite{XD}. This challenging dataset contains more videos with high abnormality ratios, as illustrated in Fig.~\ref{fig:anomalyRatio-xd}. Our batch-level selection strategy demonstrates its effectiveness in capturing potential abnormal snippets, boosting the proposed BN-WVAD to achieve an impressive AP score of 84.93\% AP when only trained on video features. Notably, our BN-WVAD outperforms the previous video-only methods~\cite{UCF,RTFM,MSL,S3R,SAS} by a large margin, even surpassing the audio-visual SOTA method MACIL-SD~\cite{MACIL-SD} by 1.53\% AP. The performance of the proposed BN-WAVD is further improved when simply concatenating audio features with video features as the input, yielding an AP score of up to 85.26\%.

\begin{table*}[]
    \centering
    \resizebox{\textwidth}{!}{%
        \begin{tabular}{ccccc|cccc|cccc}
            \myrule
            \multicolumn{5}{c|}{Module} & \multicolumn{4}{c|}{UCF-Crime} & \multicolumn{4}{c}{XD-Violence}                                                                                                                                                                             \\ \mytinyrule
            Normal Loss                 & Dropout                        & BatchNorm                       & DFM+MPP    & BLS        & AUC            & AP             & $\trm{AUC}_{abn}$ & $\trm{AP}_{abn}$ & AUC            & AP             & $\trm{AUC}_{abn}$ & $\trm{AP}_{abn}$ \\ \mytinyrule
            \checkmark                  & \checkmark                     &                                 &            &            & 65.21          & 23.79          & 55.62             & 26.11            & 61.96          & 61.54          & 55.99             & 64.94            \\
            \checkmark                  &                                & \checkmark                      &            &            & 82.97          & 25.18          & 59.40             & 28.08            & 90.74          & 72.99          & 74.13             & 74.91            \\
            \checkmark                  &                                & \checkmark                      & \checkmark &            & 86.44          & 35.94          & 70.87             & 36.67            & 94.57          & 83.33          & 83.16             & 84.60            \\
            \checkmark                  &                                & \checkmark                      & \checkmark & \checkmark & \textbf{87.24} & \textbf{36.26} & \textbf{71.71}    & \textbf{38.13}   & \textbf{94.71} & \textbf{84.93} & \textbf{83.59}    & \textbf{85.45}   \\ \myrule
        \end{tabular}%
    }
    \caption{The ablation of our proposed components on UCF-Crime~\cite{UCF} and XD-Violence~\cite{XD}. Both AUC and AP scores are reported, and $\trm{AUC}_{abn}$ and $\trm{AP}_{abn}$ denote the AUC and AP scores calculated on the subset of abnormal videos, respectively.}
    \label{tab:ablation-components}
    \vspace{-3pt}
\end{table*}

\subsection{Ablation Study}

\tbf{Effectiveness of key components.} 
To comprehensively assess the effectiveness of key components in our BN-WVAD, we report both AP and AUC scores on UCF-Crime and XD-Violence datasets by incrementally integrating each component, as presented in Table~\ref{tab:ablation-components}.
To emphasize the underestimated significance of BatchNorm in WVAD, we substitute BatchNorm with Dropout~\cite{Dropout} as an alternative to alleviate overfitting.
When solely supervised by the normal loss $\mcal{L}^\trm{nor}$, our BN-WVAD model with Dropout exhibits poor performance.
The incorporation of BatchNorm significantly improves the performance to be comparable to some existing methods~\cite{UCF,XD,GCN}, supporting our insight into the normality modeling of BatchNorm.
In this case, despite the absence of an explicit loss for abnormal videos, the gradients derived from $\mcal{L}^\trm{nor}$ are attached with knowledge of abnormal representations when back-propagating through BatchNorm~\cite{AnoOnly}, facilitating recognition of abnormal events.

The addition of the proposed DFM criterion for selection and MPP loss for optimization further enhances our BN-WVAD, making it comparable to the SOTA method UR-DMU~\cite{UR-DMU} on both datasets.
Finally, the introduction of the BLS strategy further boosts the performance of the proposed BN-WVAD to be SOTA, especially on XD-Violence, demonstrating an impressive improvement of 1.6\% AP.
Additionally, $\trm{AUC}_{abn}$ and $\trm{AP}_{abn}$, calculated on abnormal videos only, are consistently improved with the integration of each component, demonstrating the effectiveness of the proposed components in our BN-WVAD.

\begin{table}[b]
    \resizebox{\linewidth}{!}{%
        \begin{tabular}{l|cccc}
            \myrule
            Method & Criterion & Selection & UCF (AUC)      & XD (AP)        \\ \mytinyrule
            RTFM$^\dagger$~\cite{RTFM}  & FM        & SLS       & 84.11          & 74.80          \\
            RTFM$^\dagger$~\cite{RTFM}  & FM        & SBS       & 84.36          & 76.07          \\
            RTFM$^\dagger$~\cite{RTFM}  & DFM       & SLS       & 85.58          & 80.10          \\
            RTFM$^\dagger$~\cite{RTFM}  & DFM       & SBS       & \textbf{86.21}          & \textbf{82.62}          \\ \mytinyrule
            BN-WVAD (\tbf{Ours})   & FM        & SBS       & 85.84          & 81.99          \\
            BN-WVAD (\tbf{Ours})   & DFM       & SBS       & \textbf{87.24} & \textbf{84.93} \\ \myrule
        \end{tabular}%
    }
    \caption{The superiority of our DFM criterion to the widely used FM criterion~\cite{RTFM} and the applicability of DFM criterion SBS strategy. `$\dagger$' denotes the reproduced results by ourselves.}
    \label{tab:ablation-criterion}
    \vspace{-3pt}
\end{table}

\ssubsec{Applicability of DFM criterion.}
To highlight the superiority of the proposed DFM criterion compared to the widely used Feature Magnitude (FM)~\cite{RTFM}, we integrate the FM criterion and the proposed DFM criterion into RTFM~\cite{RTFM} and our BN-WAVD.
As reported in Table~\ref{tab:ablation-criterion}, regardless of adopting the SLS or SBS strategy, replacing the FM criterion with our DFM criterion in RTFM leads to a significant improvement in performance. This demonstrates the applicability of our DFM criterion to enhance existing methods.
Conversely, substituting our DFM criterion with the FM criterion in our BN-WVAD results in a notable performance decline, with a 1.4\% decrease in AUC on UCF-Crime and a 2.94\% drop in AP on XD-Violence. The performance drops further underscore the superiority of our proposed DFM criterion as the selection foundation in WVAD.

\ssubsec{Effectiveness of SBS strategy.}
The versatility of the proposed SBS strategy extends beyond our BN-WVAD, making it adaptable to other existing methods. 
As reported in Table~\ref{tab:ablation-criterion}, the introduction of the SBS strategy into RTFM~\cite{RTFM} with the FM criterion consistently enhances performance, yielding improvements of 0.25\% AUC on UCF-Crime and 1.27\% AP on XD-Violence. 
The performance gains of our SBS strategy are more remarkable when incorporated with our DFM criterion, improving the performance by 0.63\% AUC and 2.52\% AP on two datasets, respectively. The improvement divergence derived from different criteria reaffirms the efficacy of our DFM criterion in measuring the abnormality from the statistical perspective.

\begin{figure}[h]
    \centering
    \includegraphics[width=\linewidth]{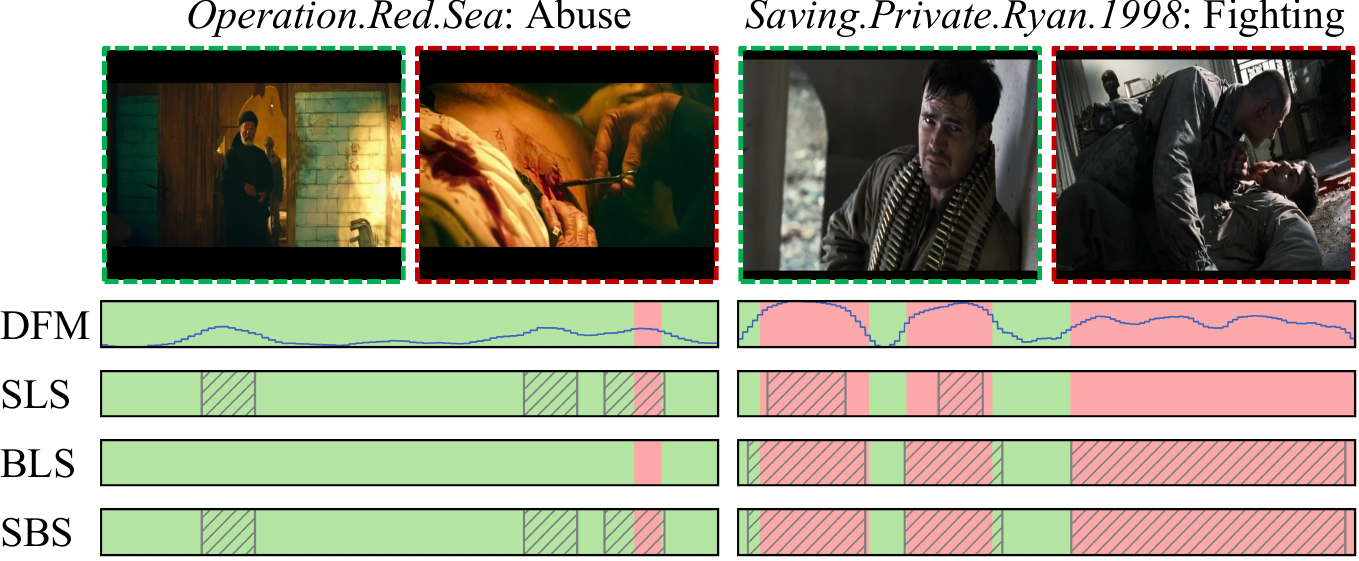}
    \caption{Visualization of DFM scores and selection results of SLS, BLS, and SBS strategies on two abnormal videos in XD-Violence~\cite{XD} with different abnormality ratios, i.e., 6.9\% and 77.7\%. Snippet-level labels are denoted by the color of the background, $\textcolor{bg-nor}{\blacksquare}$ for normal snippets and $\textcolor{bg-abn}{\blacksquare}$ for abnormal snippets.}
    \label{fig:viz-selection}
    \vspace{-3pt}
\end{figure}

The visualization in Fig.~\ref{fig:viz-selection} offers an intuitive insight into the selection results of SLS, BLS, and SBS strategies based on the DFM scores.
For clarity, we fabricate a mini-batch by selecting two abnormal videos with distinct abnormality ratios from XD-Violence.
In alignment with our earlier analysis in Sec.~\ref{sec:method-selection}, the SLS strategy exposes its limitation by choosing partial abnormal snippets in the second video with a substantial abnormality ratio.
Meanwhile, the BLS strategy struggles to identify the inconspicuous abnormal snippets in the first video.
By combining these two strategies, the proposed SBS strategy successfully mitigates the limitations of individual strategies, capturing all potential abnormal snippets in both videos.
However, our SBS strategy fails to overcome the misselection of normal snippets in the video with a low abnormality ratio, which is inevitable in WVAD with only video-level labels accessible.

\begin{figure}[t]
    \begin{subfigure}{0.45\linewidth}
        \centering
        \includegraphics[width=\linewidth]{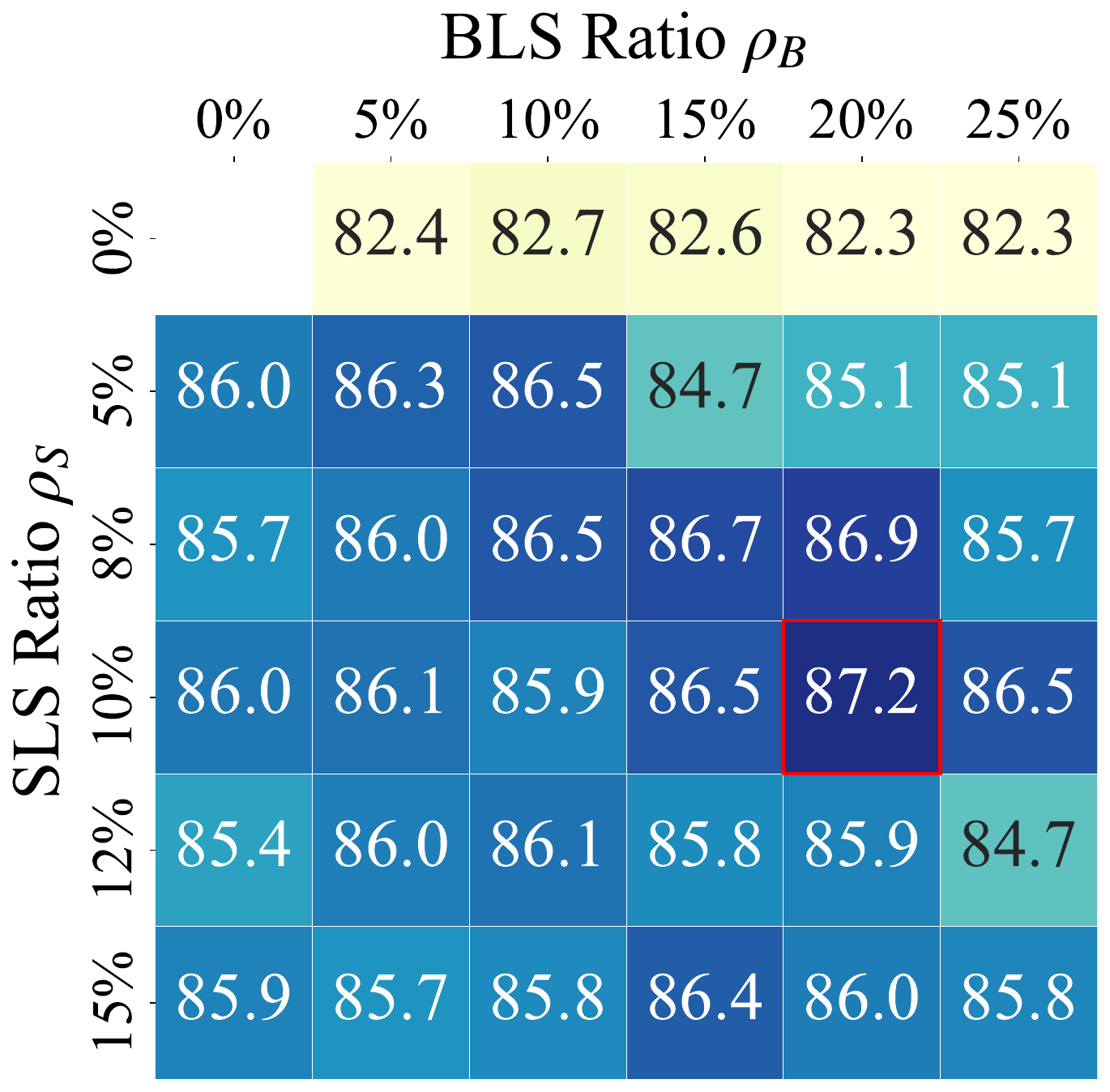}
        \caption{UCF-Crime~\cite{UCF}}
        \label{fig:ablation-rsrb-ucf}
    \end{subfigure}
    \begin{subfigure}{0.45\linewidth}
        \centering
        \includegraphics[width=\linewidth]{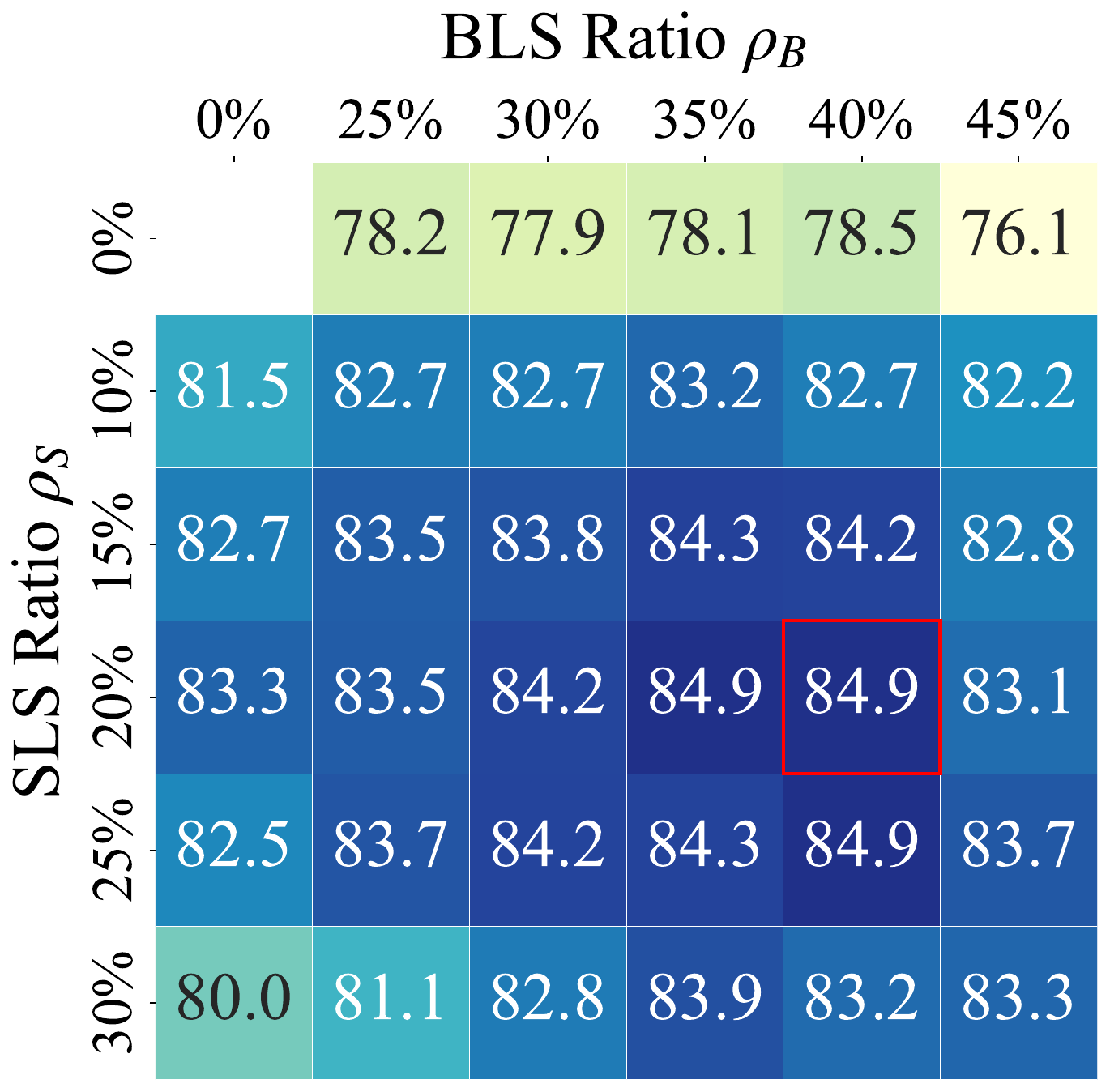}
        \caption{XD-Violence~\cite{XD}}
        \label{fig:ablation-rsrb-xd}
    \end{subfigure}
    \begin{subfigure}{0.07\linewidth}
        \vspace{0.2cm}
        \includegraphics[width=\linewidth]{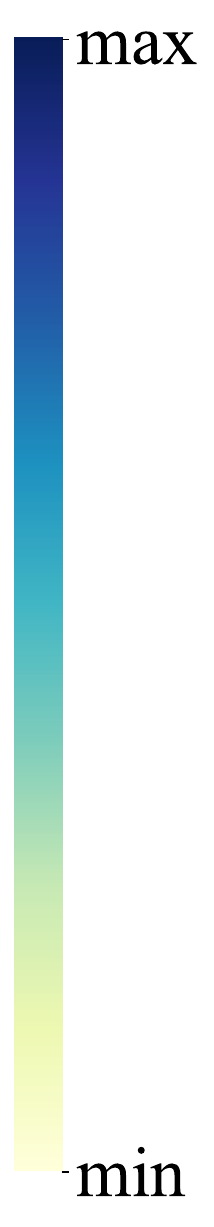}
    \end{subfigure}
    \caption{The ablation of sample-level selection ratio $\rho_s$ and batch-level selection ratio $\rho_b$ on UCF-Crime~\cite{UCF} and XD-Violence~\cite{XD}. When one of the ratios is set to 0, the corresponding selection strategy is disabled. The best results are framed by red boxes.}
    \label{fig:ablation-rsrb}
    \vspace{-3pt}
\end{figure}

\ssubsec{Ablation of selection ratios.}
We also conduct ablation studies on the selection ratios $\rho_s$ and $\rho_b$ within the SBS strategy on two datasets with distinct abnormality ratio distributions, as depicted in Fig.~\ref{fig:ablation-rsrb}.
Generally, performance improves within a certain range as both selection ratios increase, after which it experiences a decline when these ratios become excessively large.
For UCF-Crime~\cite{UCF}, optimal performance is attained when $\rho_b$=20\% and $\rho_s$=10\%, with the batch-level selection ratio closely aligning with the overall abnormality ratio in the testing set (18.2\%).
Differently, XD-Violence~\cite{XD}, characterized by a larger overall abnormality ratio (49.8\%), requires a larger selection ratio to capture potential abnormal snippets, leading to the best performance when $\rho_b$=40\% and $\rho_s$=20\%.
Despite differing optimal selection ratios, these values are relative to the overall abnormality ratio in the respective testing sets, providing instructive insights for practical application.

Notably, the individual adoption of the BLS strategy ($\rho_s$=0\%) results in a significant performance drop on both datasets.
The absence of the SLS strategy induces our BN-WAVD to be trained on the abnormal snippets from major abnormal events (e.g. Fighting).
However, it fails to address the challenging abnormal snippets associated with less frequent events such as Abuse in XD-Violence~\cite{XD}.
This observation is validated by the visualization of DFM scores for \emph{Fighting} and \emph{Abuse} in Fig.~\ref{fig:viz-selection}.

\begin{table}[t]
    \resizebox{\linewidth}{!}{%
        \begin{tabular}{ccc|ccc|ccc}
            \myrule
            \multicolumn{3}{c|}{Loss} & \multicolumn{3}{c|}{UCF-Crime (AUC)} & \multicolumn{3}{c}{XD-Violence (AP)}                                                           \\ \mytinyrule
            $\mcal{L}^{\trm{nor}}$                 & $\mcal{L}^{\trm{abn}}$                            & $\mcal{L}^{\trm{mpp}}$                           & Pred.  & DFM  & Mul.  & Pred. & DFM  & Mul.  \\ \mytinyrule
            \checkmark                &                                      &                                      & 83.0  & 81.8  & 82.9 & 73.0  & 43.5 & 68.4 \\
            \checkmark                & \checkmark                           &                                      & 82.1  & 81.3  & 81.8 & 64.6  & 34.8 & 60.4 \\
                                        &                                      & \checkmark                           & -     & 85.6  & -    & -     & 81.6    & -    \\
            \checkmark                &                                      & \checkmark                           & \textbf{86.8}  & \textbf{87.1} &  \textbf{87.2} & \textbf{83.7}  & \textbf{84.2} & \textbf{84.9} \\
            \checkmark                & \checkmark                           & \checkmark                           & 85.7  & 85.9 & 85.9 & 78.5  & 78.4  & 78.9 \\ \myrule
        \end{tabular}%
    }
    \caption{The ablation of different loss terms and anomaly score calculation strategies on UCF-Crime~\cite{UCF} and XD-Violence~\cite{XD}. `Pred.' is the prediction of anomaly classifier, and `Mul.' denotes aggregating the prediction and DFM scores by Multiplication.}
    \label{tab:ablation-loss}
    \vspace{-3pt}
\end{table}

\ssubsec{Ablation of loss terms.}
The proposed BN-WVAD resorts to two loss terms: $\mcal{L}^{\trm{nor}}$ to supervise the anomaly classifier $\mcal{C}(\cdot)$ and $\mcal{L}^{\trm{mpp}}$ to separate normal and abnormal features.
As reported in Table~\ref{tab:ablation-loss}, when solely supervised by the normal loss $\mcal{L}^{\trm{nor}}$, our DFM criterion still demonstrates considerable discrimination, achieving an AUC of 81.8\% on UCF-Crime. 
The discriminative ability of our DFM criterion is significantly boosted by incorporating the proposed MPP loss, reaching 85.6\% AUC on UCF-Crime and 81.6\% AP on XD-Violence.

\begin{figure}[h]
    \centering
    \begin{subfigure}{0.945\linewidth}
        \includegraphics[width=\linewidth]{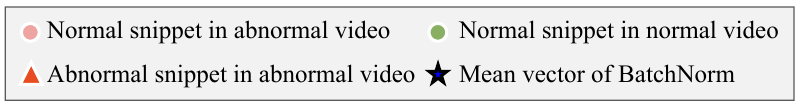}
        \vspace{-12pt}
    \end{subfigure}
    \begin{subfigure}{0.45\linewidth}
        \centering
        \includegraphics[width=\linewidth]{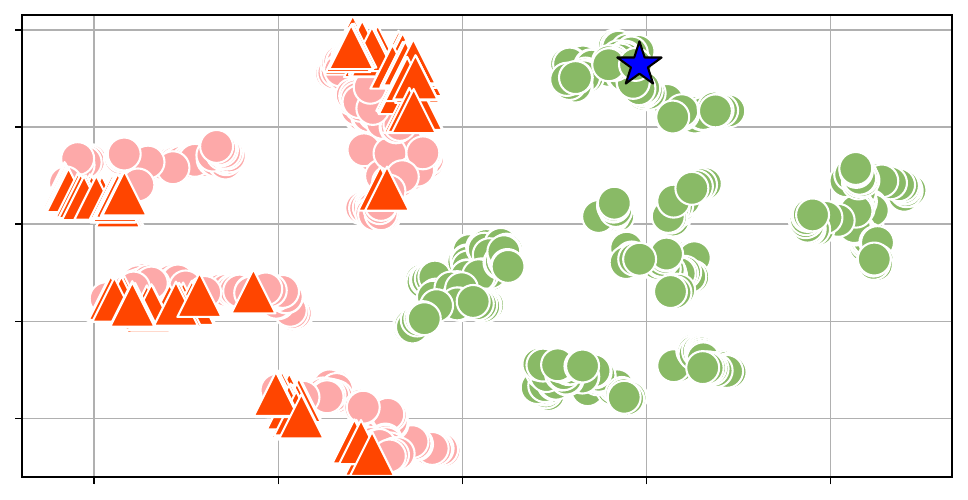}
        \caption{without MPP Loss $\mcal{L}^{\trm{mpp}}$} 
        \label{fig:viz-tsne-wompp}
    \end{subfigure}
    \hspace{0.03\linewidth}
    \begin{subfigure}{0.45\linewidth}
        \centering
        \includegraphics[width=\linewidth]{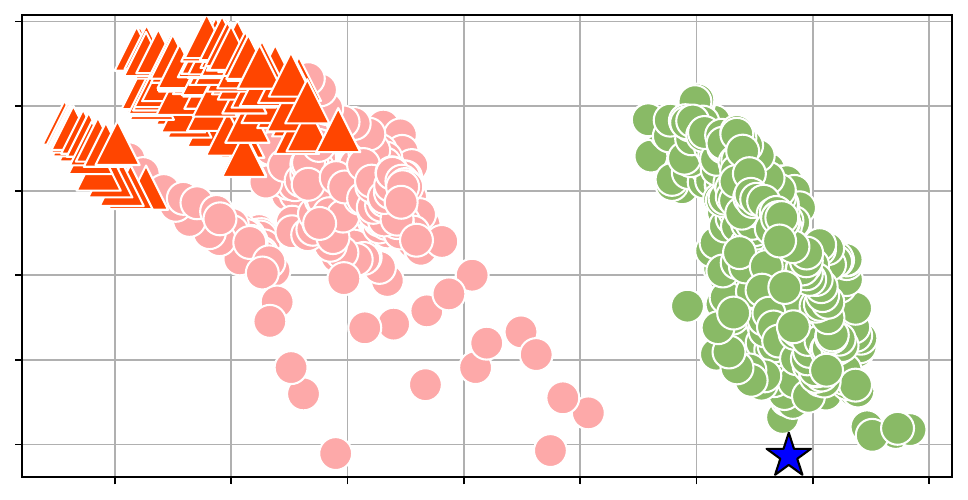}
        \caption{with MPP Loss $\mcal{L}^{\trm{mpp}}$} 
        \label{fig:viz-tsne-wmpp}
    \end{subfigure}
    
    \caption{The t-SNE visualization of hidden features with or without the supervision of MPP loss $\mcal{L}^{\trm{mpp}}$ on XD-Violence~\cite{XD}.}
    \vspace{-8pt}
    \label{fig:viz-tsne}
\end{figure}

The t-SNE visualization in Fig.~\ref{fig:viz-tsne} provides an intuitive illustration of the effectiveness of our MPP loss in enhancing feature discrimination.
The performance derived from the DFM criterion is further elevated to SOTA by incorporating $\mcal{L}^{\trm{nor}}$ and $\mcal{L}^{\trm{mpp}}$ simultaneously.
The supervision of $\mcal{L}^{\trm{nor}}$ is also beneficial to the representation learning of normality. 
The performance derived from the DFM criterion is further improved to SOTA by incorporating $\mcal{L}^{\trm{nor}}$ and $\mcal{L}^{\trm{mpp}}$ simultaneously, where the supervision of $\mcal{L}^{\trm{nor}}$ is also beneficial to the representation learning of normality.
When aggregating prediction (Pred.) and our DFM criterion, the performance of BN-WVAD is better than individual scores, demonstrating the effectiveness of our anomaly score calculation strategy.
Additionally, when incorporating the abnormal loss $\mcal{L}^{\trm{abn}}$ to supervise classifier in our BN-WVAD, the performance degrades significantly, especially on XD-Violence~\cite{XD} of 6.0\% AP decrease, even with $\mcal{L}^{\trm{mpp}}$. This observation is consistent with our earlier analysis in Sec.~\ref{sec:method-selection}, where the classifier is susceptible to label noise.

\section{Conclusion}
In this paper, we revisited the BatchNorm and introduced its statistical capacity to WVAD, presenting a novel BatchNorm-based model (BN-WVAD).
The DFM criterion was introduced to assess the abnormality of snippets, providing a statistical perspective on anomaly detection.
Moreover, we proposed an SBS strategy, inspired by BatchNorm considerations, to address the limitation within the SLS strategy.
All components introduced in our method have demonstrated effectiveness and flexibility in WVAD.

\section*{Appendix}
\appendix

Besides the experimental results reported in the main paper, we provide more experiments and analysis on our BN-WVAD in this supplementary material.
Firstly, we evaluate the proposed BN-WVAD on the other video anomaly detection dataset ShanghaiTech \cite{ShanghaiTec} with video-level labels available during training, as demonstrated in Sec. \ref{sec:suppl-shanghai}.
To further investigate the effectiveness of our BN-WAVD, we conduct more ablation studies in Sec. \ref{sec:suppl-ablation}, including the effect of different metrics in DFM calculation, the effect of momentum in BatchNorm, the effect of batch size, and comprehensive empirical analysis on the limitation of BLS.

\begin{table}[b]
    \centering
    \resizebox{0.8\columnwidth}{!}{%
    \begin{tabular}{lccc}
    \toprule
    Method              & Venue        & Feature & AUC (\%)       \\ \midrule
    Sultani et al.~\cite{UR-DMU}      & CVPR$'$18     & C3D     & 86.30          \\
    GCN~\cite{GCN}             & CVPR$'$19     & TSN     & 84.44          \\
    CLAWS~\cite{CLAWS}              & ECCV$'$20     & C3D     & 89.67          \\
    MIST~\cite{MIST}               & CVPR$'$21     & I3D     & 94.83          \\
    RTFM~\cite{RTFM}               & ICCV$'$21     & I3D     & 97.32          \\
    MSL~\cite{MSL}                 & AAAI$'$22     & I3D     & 97.32          \\
    S3R~\cite{S3R}             & ECCV$'$22     & I3D     & 97.48          \\
    UR-DMU$^\dagger$~\cite{UR-DMU}& AAAI$'$23 & I3D    & 96.90 \\ \midrule
    \multicolumn{2}{c}{BN-WVAD (\textbf{Ours})} & I3D     & \textbf{97.61} \\ \bottomrule
    \end{tabular}%
    }
    \caption{Comparison of AUC (\%) on ShanghaiTech~\cite{ShanghaiTec}. `$\dagger$' denotes the reproduction of open-source code~\cite{UR-DMU} by ourselves.}
    \label{tab:suppl-comparison-sh}
\end{table}

\section{Comparison on ShanghaiTech} \label{sec:suppl-shanghai}
\begin{figure}[t]
    \begin{subfigure}{0.55\linewidth}
        \centering
        \includegraphics[width=\linewidth]{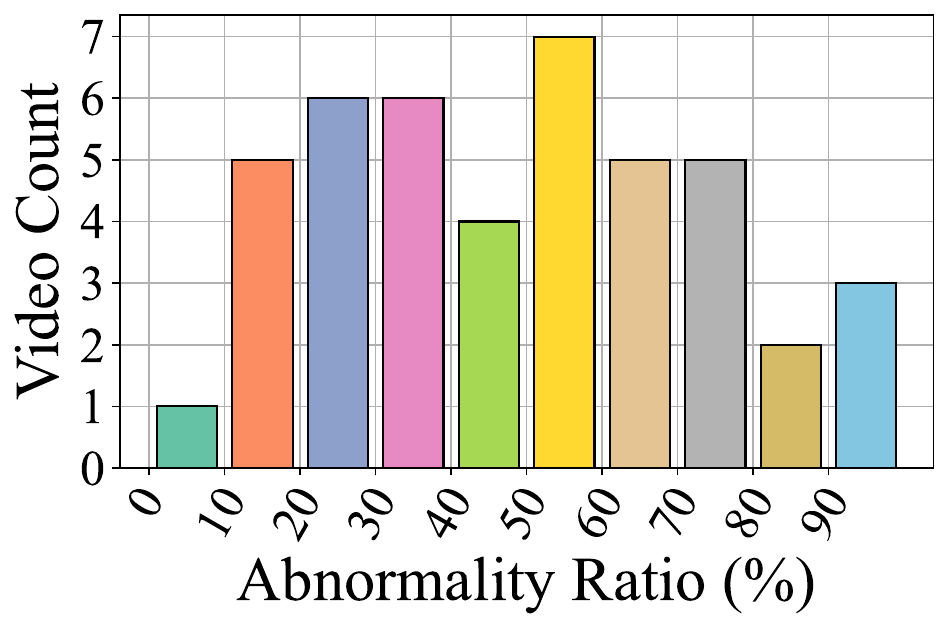}
        \caption{Abnormality ratio distribution}
        \label{fig:suppl-ShanghaiTech-abnormalRatio}
    \end{subfigure}
    \begin{subfigure}{0.37\linewidth}
        \centering
        \includegraphics[width=\linewidth]{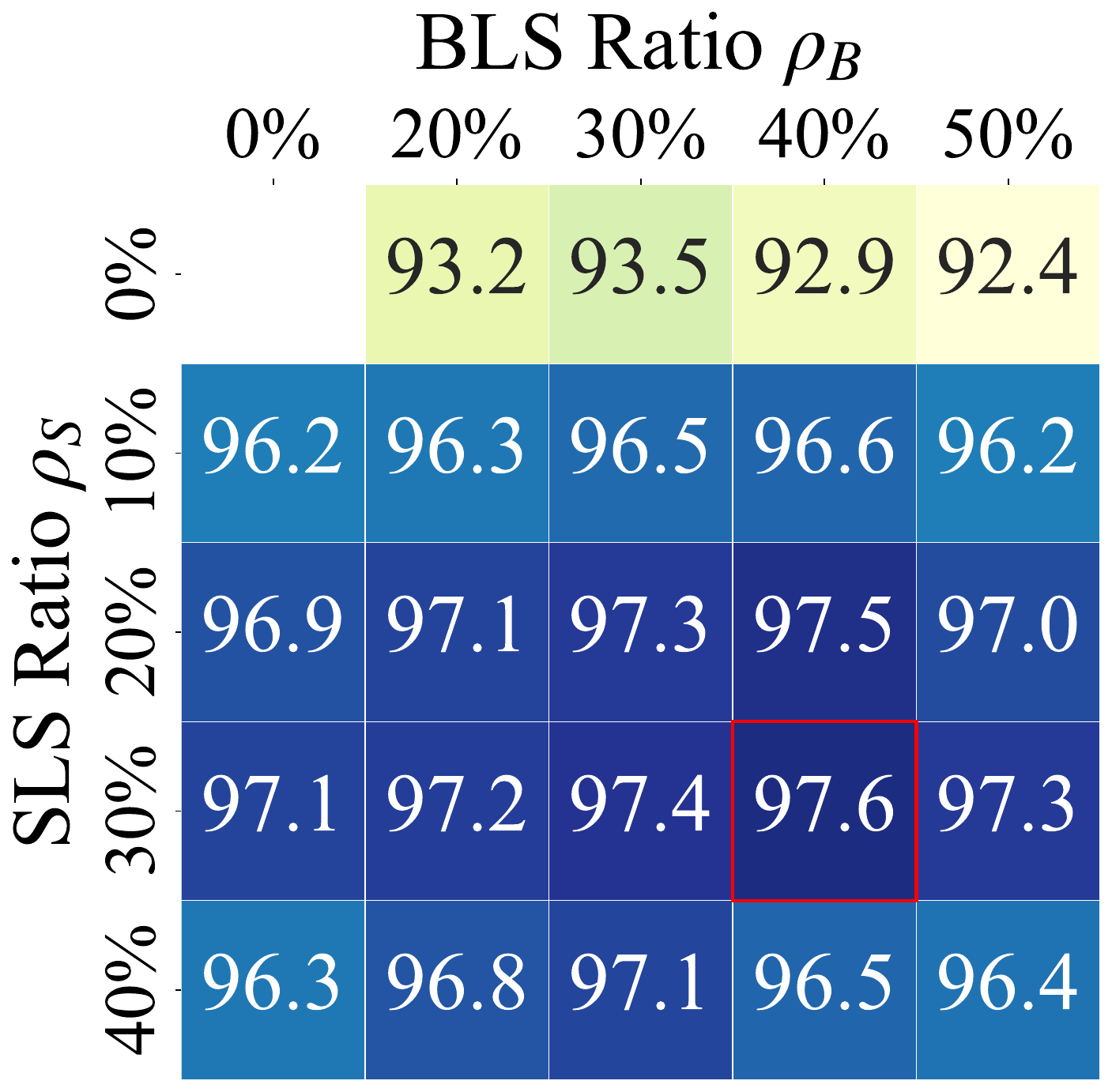}
        \caption{Ablation of selection ratios}
        \label{fig:suppl-ShanghaiTech-rsrb}
    \end{subfigure}
    \begin{subfigure}{0.06\linewidth}
        \vspace{0.2cm}
        \includegraphics[width=\linewidth]{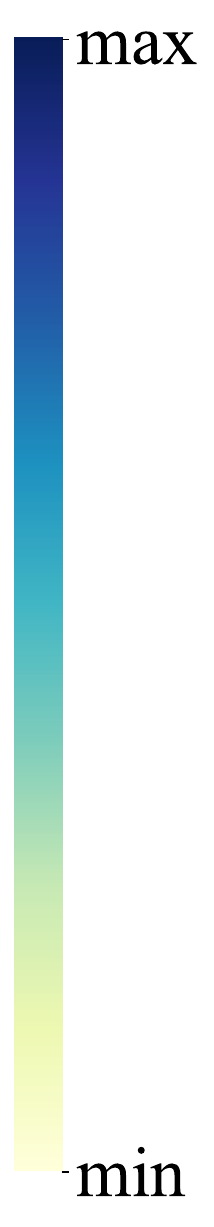}
    \end{subfigure}
    \caption{(a) The illustration of abnormality ratio distribution on ShanghaiTech~\cite{ShanghaiTec}. (b) The ablation of sample-level selection ratio $\rho_s$ and batch-level selection ratio $\rho_b$ on ShanghaiTech~\cite{ShanghaiTec}. When one of the ratios is set to 0, the corresponding selection strategy is disabled. The best results are framed by red boxes.}
    \label{fig:suppl-ShanghaiTech}
    \vspace{-3pt}
\end{figure}

\emph{ShanghaiTech}~\cite{ShanghaiTec} is a medium-scale video anomaly detection dataset compared with UCF-Crime~\cite{UCF} and XD-Violence~\cite{XD}. It collects 437 videos from fixed-angle street video surveillance, including 307 normal videos and 130 anomaly videos. This dataset was initially published targeting unsupervised video anomaly detection, where only normal videos are accessible during training. Zhong et al.~\cite{GCN} reorganized the dataset by introducing a subset of anomaly videos into the training set, satisfying the weakly supervised setting. The video-level labels are available during training, while the frame-level labels are not provided. Specifically, 238 videos are used for training and 199 videos are used for testing. Both training and testing sets contain all 13 abnormal classes. We leverage AUC as the evaluation metric following~\cite{GCN} and compare our BN-WVAD with previous methods~\cite{UCF,GCN,CLAWS,MIST,RTFM,MSL,S3R}.
In particular, the sample-level selection ratio $\rho_s$ and batch-level ratio $\rho_b$ are set to 0.3 and 0.4, respectively, positively correlated with the abnormality ratio of the ShanghaiTech dataset, i.e., 46.6\%, as presented in Fig.~\ref{fig:suppl-ShanghaiTech-abnormalRatio}.
With these selection ratio settings, our BN-WVAD achieves the best performance on ShanghaiTech~\cite{ShanghaiTec}, as illustrated in Fig~\ref{fig:suppl-ShanghaiTech-rsrb}.

We report the empirical comparison on ShanghaiTech~\cite{ShanghaiTec} in Table~\ref{tab:suppl-comparison-sh}. Consistently, our BN-WVAD outperforms previous methods~\cite{UCF,GCN,CLAWS,MIST,RTFM,MSL,S3R,UR-DMU}, demonstrating the effectiveness and generalization of our proposed method in the weakly supervised setting. Although the performance gap between our BN-WVAD and the previous SOTA method S3R~\cite{S3R} is not as significant as that on UCF-Crime~\cite{UCF} and XD-Violence~\cite{XD}, our BN-WVAD still achieves the best performance without fine-tuning the hyper-parameters on ShanghaiTech~\cite{ShanghaiTec}. On the other hand, due to the limited number of training data in ShanghaiTech~\cite{ShanghaiTec}, the statistics captured by BatchNorm in our BN-WVAD are prone to overfitting to the training data, leading to a performance drop compared to the results on the other two large-scale datasets~\cite{XD,UCF}.

\section{More Ablation Studies} \label{sec:suppl-ablation}

In this section, we provide more ablation studies on the proposed BN-WVAD. Specifically, we investigate the effect of different metrics within DFM calculation, the effect of varying momentum settings in BatchNorm, and the effect of different batch size settings.
Additionally, we comprehensively analyze the limitation of BLS in our BN-WVAD by reporting the AP of each abnormal class on XD-Violence.

\subsection{Different Metric of DFM calculation} \label{sec:metric}
Besides the Mahalanobis distance \cite{de2000mahalanobis} used in the main paper, we also investigate the effect of other metrics in DFM calculation, including common Euclidean distance and cosine similarity. The results are reported in Table~\ref{tab:suppl-metric}.

We can observe that the Mahalanobis distance achieves the best performance, which is consistent with the results reported in the main paper. 
When employing the Euclidean distance, the performance is slightly worse than the Mahalanobis distance, which is because the Euclidean distance is a special case of the Mahalanobis distance under the assumption that the Gaussian distributions of different features are independent and scale-invariant, sharing the same variance of 1.
The Cosine Similarity performs the worst compared to the other two metrics, with an AUC of 85.33\% on UCF-Crime~\cite{UCF} and an AP of 81.82\% on XD-Violence~\cite{XD}.
We conjecture the inferior performance derived from the Cosine Similarity is because the magnitude of the feature vectors is not considered in the calculation.
However, the divergence of feature magnitude is also significant in distinguishing abnormal snippets from normal snippets
, which even independently serves as an abnormality criterion in RTFM \cite{RTFM}.

\begin{table}[t]
    \centering
    \resizebox{\columnwidth}{!}{%
    \begin{tabular}{l|cc}
        \myrule
    Metric               & UCF-Crime (\%) & XD-Violence (\%) \\ \mytinyrule
    Cosine Similarity    & 85.33          & 81.82            \\
    Euclidean Distance   & 86.51          & 83.45            \\
    Mahalanobis Distance & \tbf{87.24}          & \tbf{84.93}            \\ \myrule
    \end{tabular}%
    }
    \caption{The ablation of different metrics in DFM calculation on UCF-Crime~\cite{UCF} and XD-Violence~\cite{XD}. AUC and AP scores are reported on UCF-Crime~\cite{UCF} and XD-Violence~\cite{UCF}, respectively.}
    \label{tab:suppl-metric}
    \end{table}

\subsection{The Effect of Momentum in BatchNorm} \label{sec:momentum}
The momentum in BatchNorm is a hyper-parameter that controls the contribution of the current batch statistics to the running mean and variance, which works as an exponential moving average (EMA) update as follows:
\begin{align}
    \sbf{\hat{\mu}}      & = (1-\alpha) \sbf{\hat{\mu}} + \alpha \sbf{\mu},           \\
    \sbf{\hat{\sigma}}^2 & = (1-\alpha) \sbf{\hat{\sigma}}^2 + \alpha \sbf{\sigma}^2, 
\end{align}
where $\sbf{\hat{\mu}}$ and $\sbf{\hat{\sigma}}^2$ are the running mean and variance, respectively, and $\sbf{\mu}$ and $\sbf{\sigma}^2$ are the mean and variance of the current batch, respectively. The momentum $\alpha$ is set to 0.1 by default in PyTorch~\cite{paszke2017automatic}, which is also adopted in the proposed BN-WVAD.

In this section, we investigate the effect of different momentum $\alpha$ settings in BatchNorm. The results are reported in Table~\ref{tab:suppl-momentum} with momentum $\alpha\in\{0.01, 0.1, 0.2, 0.5, 1\}$. When $\alpha$ is set to 0.01, the performance is slightly worse than the default setting of $\alpha$=0.1, which is because the running mean and variance are updated too sluggishly to capture the normality representation of the current mini-batch. Increasing the momentum $\alpha$ to be larger than 0.1, the performance drops gradually, especially when $\alpha$ is set to 1, the performance degrades
significantly to 81.64\% AUC on UCF-Crime~\cite{UCF} and 68.69\% AP on XD-Violence~\cite{XD}.
In this specific case of $\alpha$=1, the running mean $\sbf{\hat{\mu}}$ and variance $\sbf{\hat{\sigma}}^2$ are not updated at all, where the statistics of each mini-batch are used to normalize the features of the whole training process.
The absence of the EMA update in BatchNorm leads to the overfitting to the training data of each mini-batch, resulting in a dramatic performance drop. This observation is consistent with our earlier analysis in the main text, motivating the introduction of the momentum in BatchNorm.

\begin{table}[t]
    \centering
    \resizebox{0.85\columnwidth}{!}{%
    \begin{tabular}{ccc}
    \myrule
    Momentum $\alpha$ & UCF-Crime (\%) & XD-Violence (\%) \\ \mytinyrule
    0.01     & 87.01                 & 83.48                \\
    \textbf{0.1} (\textbf{Ours})      & \textbf{87.24}                 & \textbf{84.93}                  \\
    0.2      & 86.98                 &  84.61             \\
    0.5      & 84.69                 & 84.52                 \\
    1        & 81.64                 & 68.69                 \\ \myrule
    \end{tabular}%
    }
    \caption{The ablation of different momentum $\alpha\in\{0.01, 0.1, 0.2,\\ 0.5, 1\}$ settings in BatchNorm on UCF-Crime~\cite{UCF} and XD-Violence~\cite{XD}. AUC and AP scores are reported on UCF-Crime~\cite{UCF} and XD-Violence~\cite{UCF}, respectively.}
    \label{tab:suppl-momentum}
    \end{table}

\subsection{The Effect of Batch Size} \label{sec:batchsize}
The essential motivation of the proposed BN-WVAD is to leverage the statistics captured by BatchNorm to distinguish abnormal snippets from normal snippets. Despite the EMA update in BatchNorm to capture the normality representation of the whole training set, the statistics of each mini-batch still play a crucial role in the training process. On the one hand, the statistics of each mini-batch are used to normalize the features of the whole training process. On the other hand, the ratios of normal and abnormal input videos within each mini-batch also determine the statistics captured by BatchNorm. Therefore, we investigate the effect of different batch size settings on the performance of our BN-WVAD by varying the batch size of normal and abnormal videos in each mini-batch, respectively.

\begin{figure}[h]
    \begin{subfigure}{0.44\linewidth}
        \centering
        \includegraphics[width=0.9\linewidth]{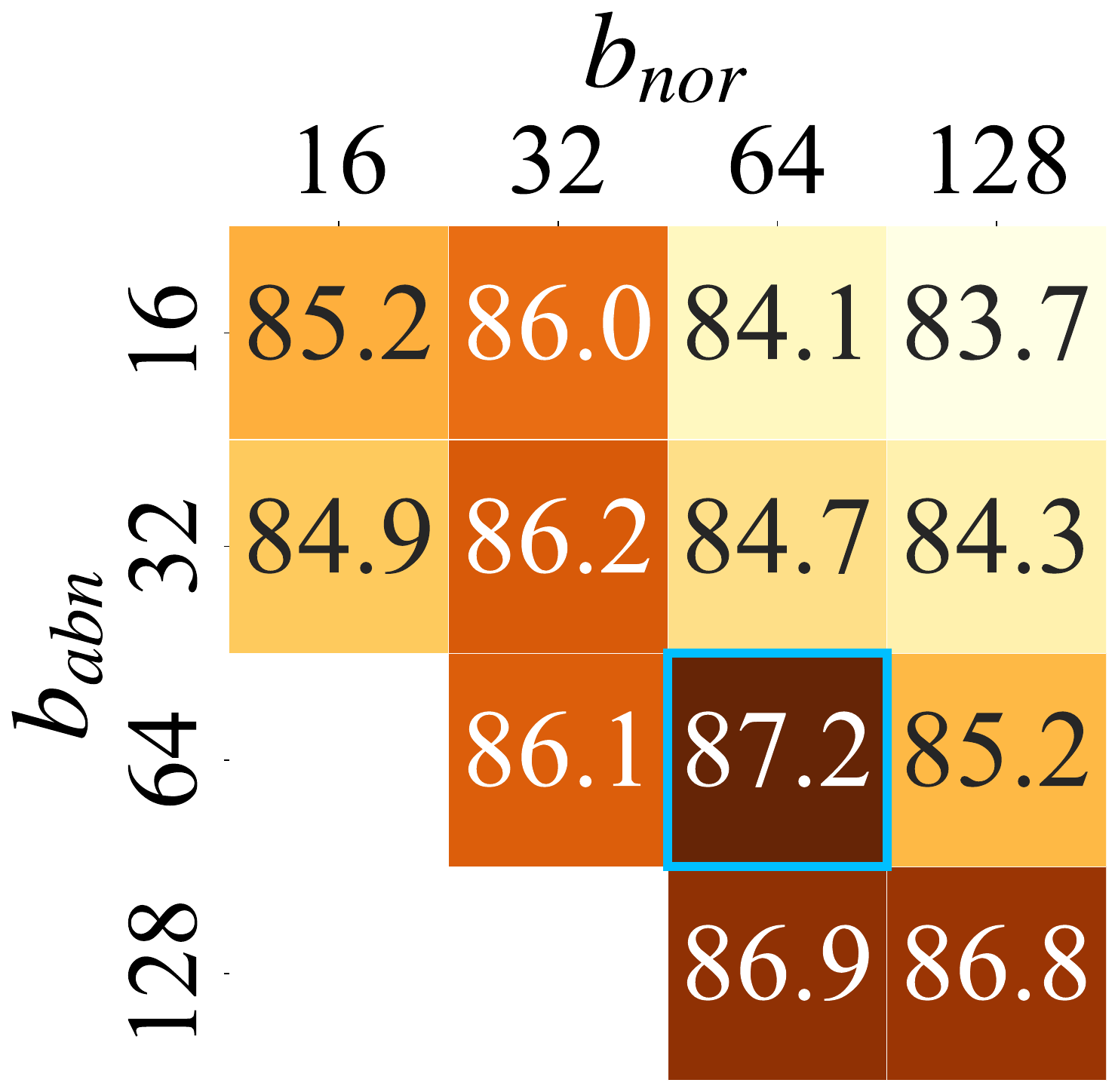}
        \caption{UCF-Crime~\cite{UCF}}
        \label{fig:suppl-batch-size-ucf}
    \end{subfigure}
    \begin{subfigure}{0.44\linewidth}
        \centering
        \includegraphics[width=0.9\linewidth]{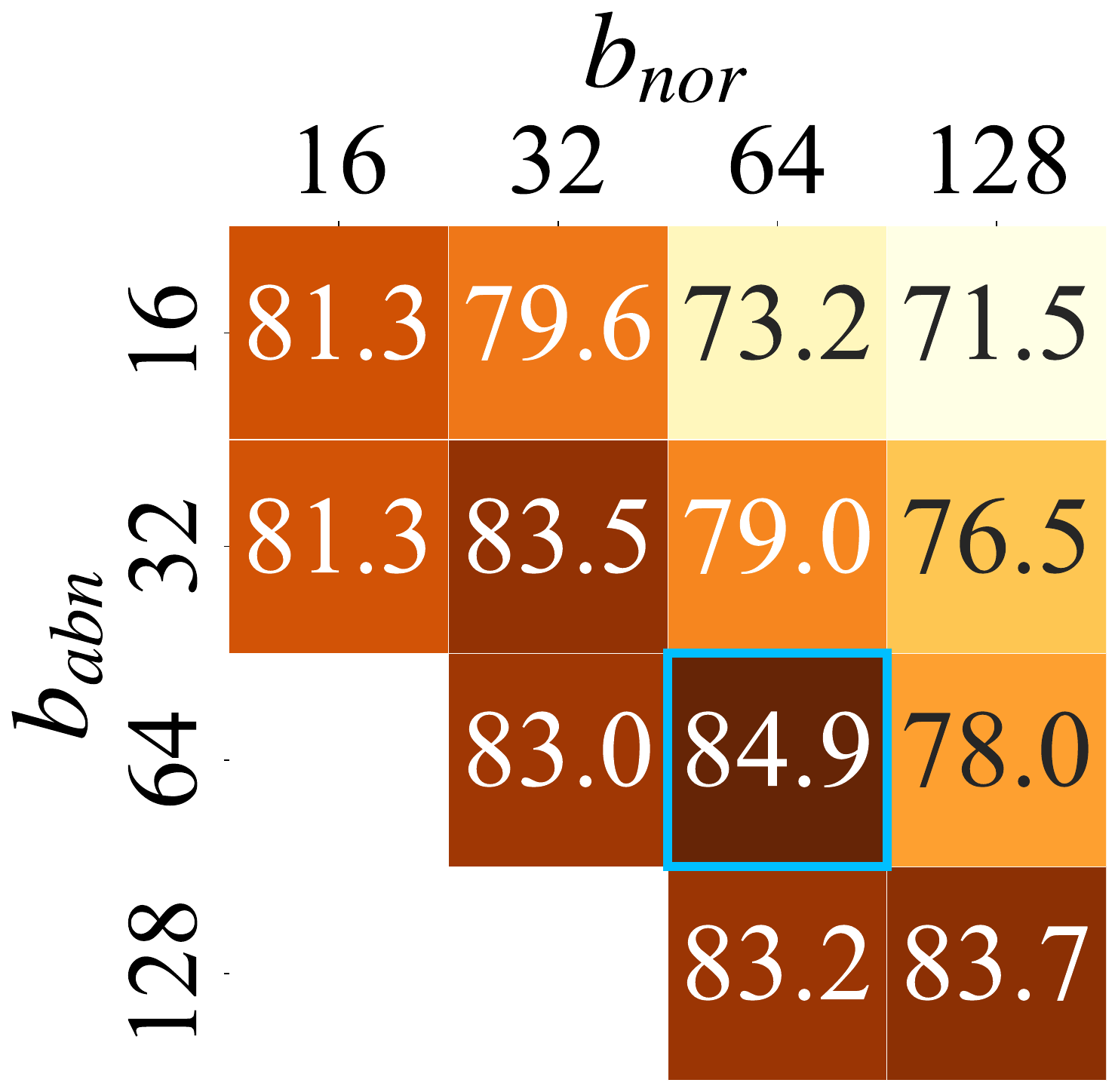}
        \caption{XD-Violence~\cite{XD}}
        \label{fig:suppl-batch-size-xd}
    \end{subfigure}
    \begin{subfigure}{0.08\linewidth}
        \vspace{0.22cm}
        \includegraphics[width=\linewidth]{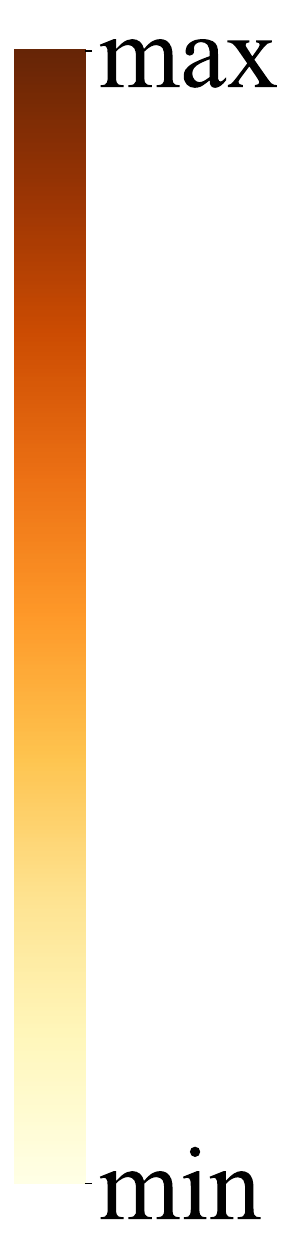}
    \end{subfigure}
    \caption{The illustration of batch size settings on UCF-Crime~\cite{UCF} and XD-Violence~\cite{XD}. The batch size of normal videos $b_\trm{nor}$ and the batch size of abnormal videos $b_\trm{abn}$ are set to be 16, 32, 64, and 128, respectively. The best results are framed by blue boxes.}
    \label{fig:suppl-batch-size}
    \vspace{-3pt}
\end{figure}
    
\begin{figure*}[t]
    \centering
    \begin{subfigure}{0.162\textwidth}
        \centering
        \includegraphics[width=\linewidth]{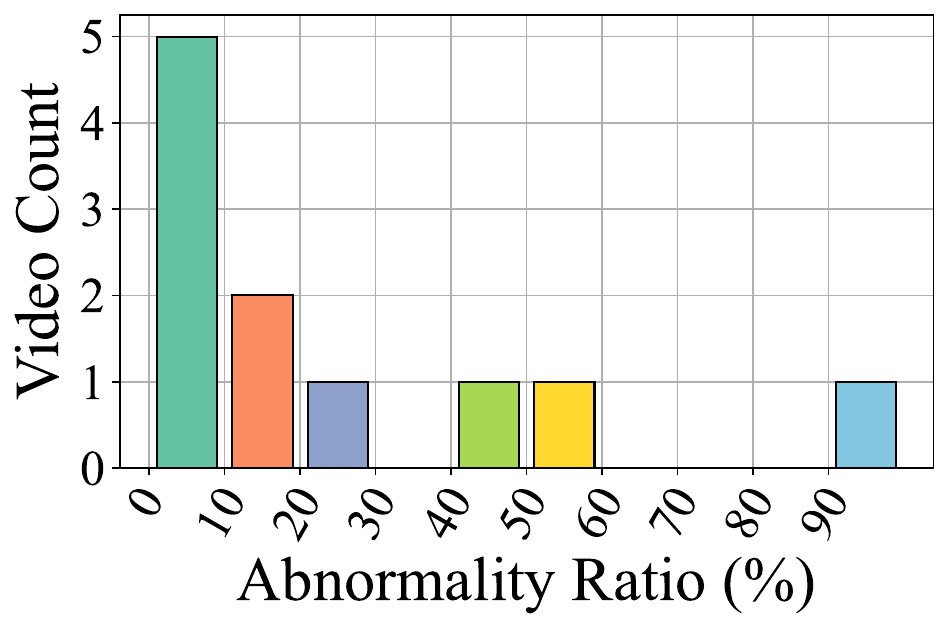}
        \caption{Abuse}
    \end{subfigure}
    \begin{subfigure}{0.162\textwidth}
        \centering
        \includegraphics[width=\linewidth]{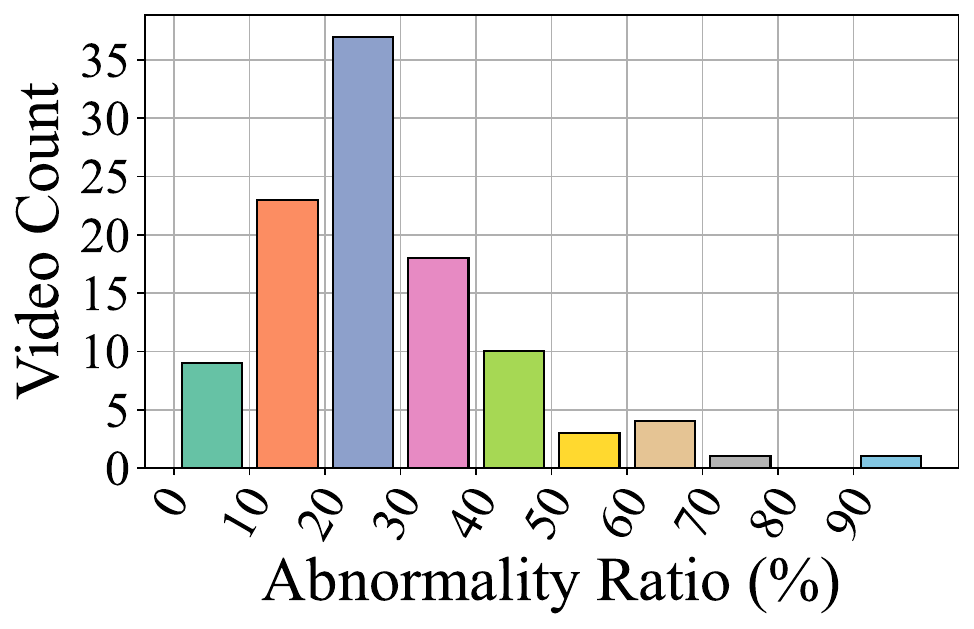}
        \caption{Car Accident}
    \end{subfigure}
    \begin{subfigure}{0.162\textwidth}
        \centering
        \includegraphics[width=\linewidth]{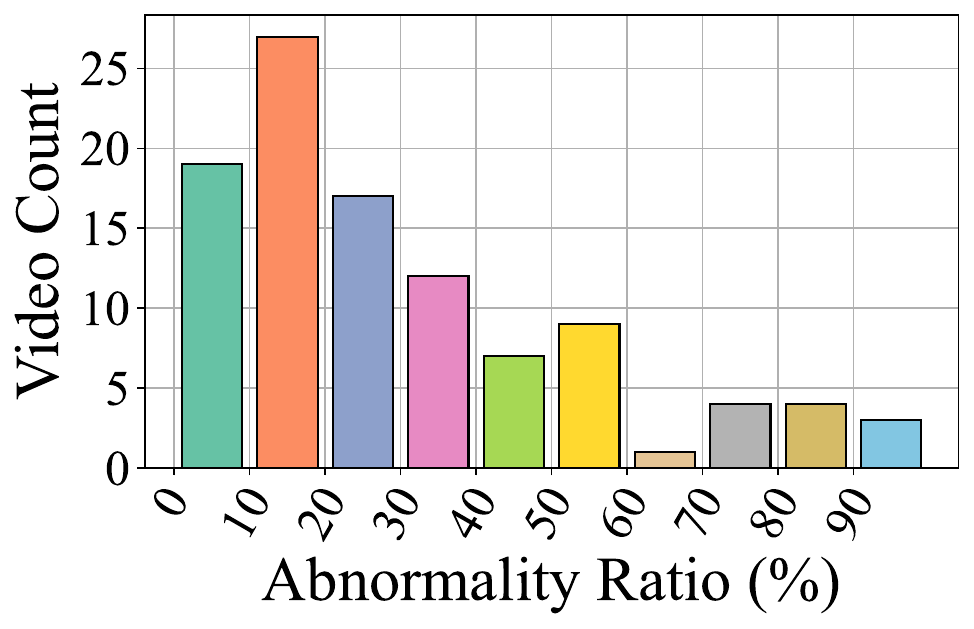}
        \caption{Explosion}
    \end{subfigure}
    \begin{subfigure}{0.162\textwidth}
        \centering
        \includegraphics[width=\linewidth]{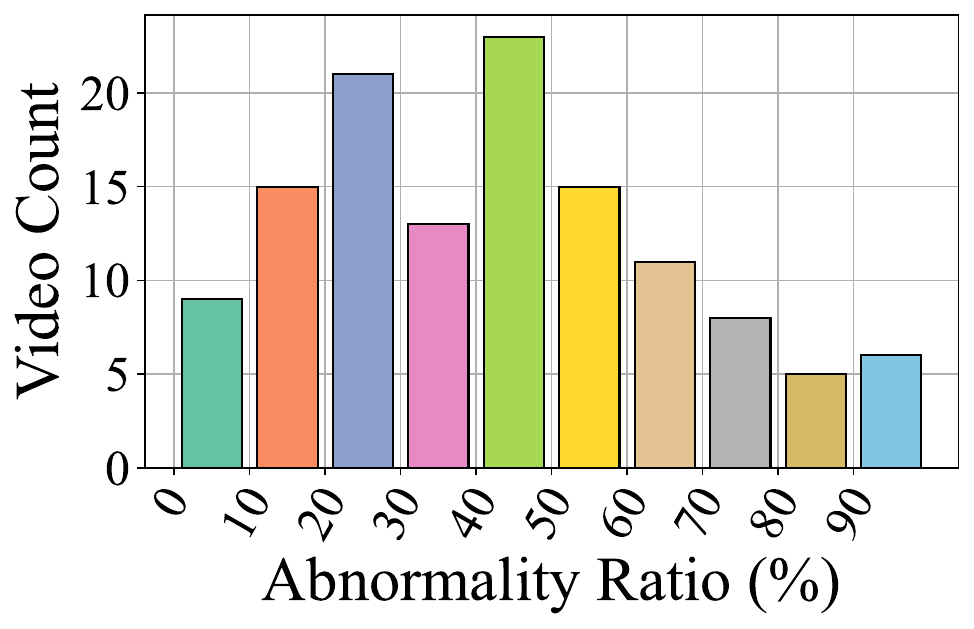}
        \caption{Fighting}
    \end{subfigure}
    \begin{subfigure}{0.162\textwidth}
        \centering
        \includegraphics[width=\linewidth]{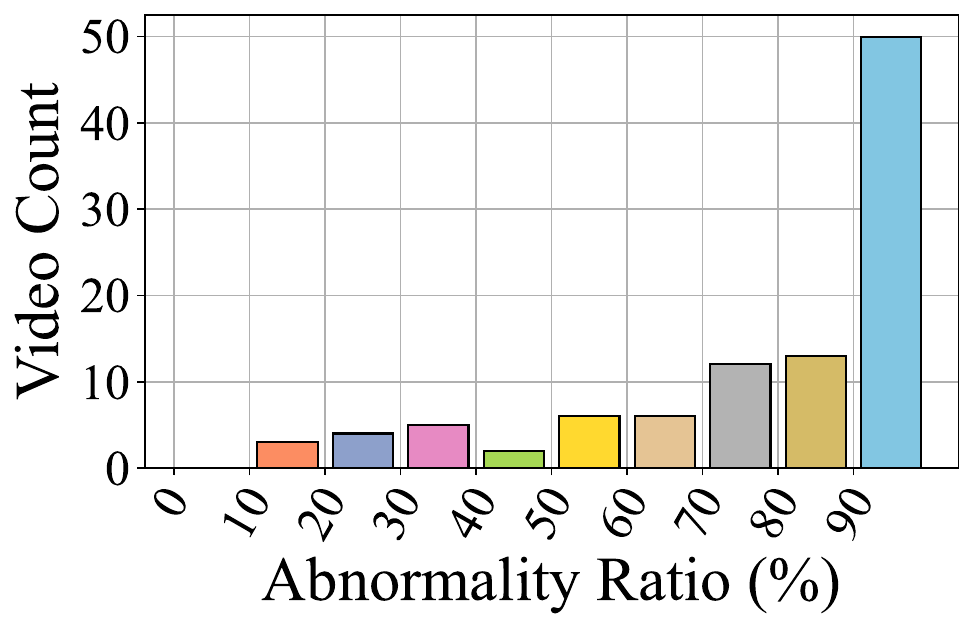}
        \caption{Riot}
    \end{subfigure}
    \begin{subfigure}{0.162\textwidth}
        \centering
        \includegraphics[width=\linewidth]{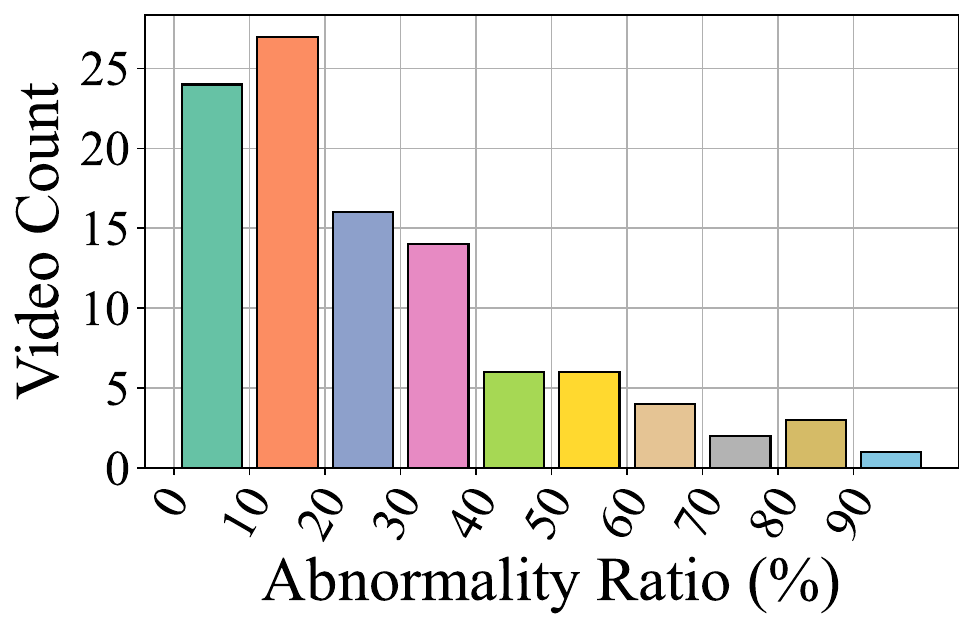}
        \caption{Shooting}
    \end{subfigure}
    \caption{The illustration of abnormality ratio distribution of different abnormal classes on XD-Violence~\cite{XD}.}
    \label{fig:suppl-abnormality-ratio}
\end{figure*}

The results are reported in Fig.~\ref{fig:suppl-batch-size}, where the batch size of normal videos $b_\trm{nor}$ and the batch size of abnormal videos $b_\trm{abn}$ are set to be 16, 32, 64, and 128, respectively. 
We only vary the batch size of normal videos and abnormal videos, while keeping other training hyper-parameters fixed to the default settings in the main paper.
In particular, due to the demand for pairwise MPP loss calculation, $b_\trm{abn}$ can only be set to 16, 32 when $b_\trm{nor}$=16, and 16, 32, 64 when $b_\trm{nor}$=32, respectively.
We can observe that the performance on both UCF-Crime~\cite{UCF} and XD-Violence~\cite{XD} achieve the best when $b_\trm{nor}$ and $b_\trm{abn}$ are both set to be 64.
This optimal batch size setting is consistent with the default settings of UR-DMU~\cite{UR-DMU}, where our BN-WVAD implementation is heavily based.

When concurrently changing $b_\trm{nor}$ and $b_\trm{abn}$ to be smaller than 64, i.e., 32 and 16, the performance degrades significantly, which is because the statistics captured by BatchNorm are partial and prone to overfitting to the training data of each mini-batch.
When $b_\trm{nor}$ and $b_\trm{abn}$ are enlarged to be 128, the performance on both datasets also slightly drops, which is because the statistics captured by BatchNorm are diluted by the enlarged batch size, leading to a less discriminative representation of normality.
Furthermore, when $b_\trm{nor}$ is larger than $b_\trm{abn}$, the captured statistics are dominated by the normal videos, motivating a more discriminative representation of normality. However, the training focus on the abnormal snippets is reduced, leading to a performance drop. We infer that tuning the training weight on normal and abnormal snippets may mitigate this issue and achieve better performance.
On the other hand, when $b_\trm{nor}$ is smaller than $b_\trm{abn}$, the performance is even worse than the case when $b_\trm{nor}$ is larger than $b_\trm{abn}$. We conjecture the main reason is that the statistics computed by BatchNorm are distracted by multiple abnormal snippets, failing to capture a prototypical representation of normality.

\subsection{The Limitation of BLS Strategy} \label{sec:bls}
To comprehensively analyze the limitation of BLS strategy in our BN-WVAD, we report the AP of each abnormal class on XD-Violence \cite{XD} in Table~\ref{tab:selection-limitation}. The sample-level selection ratio $\rho_s$ and batch-level selection ratio $\rho_b$ are set to 0.2 and 0.4, respectively.
We can observe that individually adopting the BLS strategy performs worse than the SLS strategy in all abnormal classes, especially for the abnormal classes with low abnormality ratios, such as Abuse, Car Accident, and Explosion, as illustrated in Fig.~\ref{fig:suppl-abnormality-ratio}.
This observation is consistent with our earlier analysis in the main text, where the BLS strategy may overlook the inconspicuous abnormal snippets in videos with a low abnormality ratio.
Notably, the SLS strategy even performs better than the BLS strategy on the abnormal class, \emph{Riot}, which is characterized by a high abnormality ratio. We conjecture the main reason for this counterintuitive observation is that this specific abnormal class is long-lasting but stationary, short snippets are sufficient to capture the abnormality.

When combining the SLS and BLS strategies, the proposed SBS strategy mitigates the limitations of individual strategies, achieving the best overall performance.
However, on the abnormal classes, \emph{Abuse} and \emph{Explosion}, the SBS strategy is still inferior to the SLS strategy, which is because the incorporation of the BLS strategy reduces the training focus on the abnormal snippets from these two abnormal classes, whose abnormality ratios are relatively low as illustrated in Fig.~\ref{fig:suppl-abnormality-ratio}.

\begin{table}[]
    \centering
    \resizebox{\columnwidth}{!}{%
    \begin{tabular}{lccccccc}
    \toprule
    Selection & Abu.          & C.A.           & Expl.          & Figt.         & Riot           & Shoot          & All            \\ \midrule
    SLS       & \textbf{43.50} & 37.48          & \textbf{55.71} & 79.15         & 95.69          & 57.84          & 83.55          \\
    BLS       & 30.71         & 33.28          & 51.83          & 75.09         & 92.17          & 56.69          & 78.55          \\
    SBS       & 41.90         & \textbf{39.18} & 54.74          & \textbf{84.90} & \textbf{96.18} & \textbf{58.52} & \textbf{84.93} \\ \bottomrule
    \end{tabular}%
    }
    \caption{Class-wise AP (\%) of different selection strategies on XD-Violence~\cite{XD}. `Abu.' denotes Abuse, `C.A.' denotes Car Accident, `Expl.' denotes Explosion, `Figt.' denotes Fighting, and `Shoot' denotes Shooting.}
    \label{tab:selection-limitation}
    \end{table}

{
    \small
    \bibliographystyle{ieeenat_fullname}
    \bibliography{main}

\begin{thebibliography}{51}
\providecommand{\natexlab}[1]{#1}
\providecommand{\url}[1]{\texttt{#1}}
\expandafter\ifx\csname urlstyle\endcsname\relax
  \providecommand{\doi}[1]{doi: #1}\else
  \providecommand{\doi}{doi: \begingroup \urlstyle{rm}\Url}\fi

\bibitem[Abadi et~al.(2016)Abadi, Barham, Chen, Chen, Davis, Dean, Devin,
  Ghemawat, Irving, Isard, et~al.]{abadi2016tensorflow}
Mart{\'\i}n Abadi, Paul Barham, Jianmin Chen, Zhifeng Chen, Andy Davis, Jeffrey
  Dean, Matthieu Devin, Sanjay Ghemawat, Geoffrey Irving, Michael Isard, et~al.
\newblock Tensorflow: a system for large-scale machine learning.
\newblock In \emph{OSDI}, pages 265--283, 2016.

\bibitem[Abati et~al.(2019)Abati, Porrello, Calderara, and
  Cucchiara]{abati2019latent}
Davide Abati, Angelo Porrello, Simone Calderara, and Rita Cucchiara.
\newblock Latent space autoregression for novelty detection.
\newblock In \emph{CVPR}, pages 481--490, 2019.

\bibitem[Andrews et~al.(2002)Andrews, Tsochantaridis, and Hofmann]{MIL}
Stuart Andrews, Ioannis Tsochantaridis, and Thomas Hofmann.
\newblock Support vector machines for multiple-instance learning.
\newblock \emph{NeurIPS}, 15, 2002.

\bibitem[Carreira and Zisserman(2017)]{I3D}
Joao Carreira and Andrew Zisserman.
\newblock Quo vadis, action recognition? a new model and the kinetics dataset.
\newblock In \emph{CVPR}, pages 4727--4733, 2017.

\bibitem[De~Maesschalck et~al.(2000)De~Maesschalck, Jouan-Rimbaud, and
  Massart]{de2000mahalanobis}
Roy De~Maesschalck, Delphine Jouan-Rimbaud, and D{\'e}sir{\'e}~L Massart.
\newblock The mahalanobis distance.
\newblock \emph{Chemometrics and Intelligent Laboratory Systems}, 50\penalty0
  (1):\penalty0 1--18, 2000.

\bibitem[Di~Biase et~al.(2021)Di~Biase, Blum, Siegwart, and
  Cadena]{di2021pixel}
Giancarlo Di~Biase, Hermann Blum, Roland Siegwart, and Cesar Cadena.
\newblock Pixel-wise anomaly detection in complex driving scenes.
\newblock In \emph{CVPR}, pages 16918--16927, 2021.

\bibitem[Fan et~al.(2023)Fan, Yu, Lu, and Han]{SAS}
Yidan Fan, Yongxin Yu, Wenhuan Lu, and Yahong Han.
\newblock Weakly-supervised video anomaly detection with snippet anomalous
  attention.
\newblock \emph{arXiv preprint arXiv:2309.16309}, 2023.

\bibitem[Feng et~al.(2021)Feng, Hong, and Zheng]{MIST}
JiaChang Feng, FaTing Hong, and WeiShi Zheng.
\newblock Mist: Multiple instance self-training framework for video anomaly
  detection.
\newblock In \emph{CVPR}, pages 14009--14018, 2021.

\bibitem[Gardner~Jr(1985)]{gardner1985exponential}
Everette~S Gardner~Jr.
\newblock Exponential smoothing: The state of the art.
\newblock \emph{Journal of forecasting}, 4\penalty0 (1):\penalty0 1--28, 1985.

\bibitem[Gemmeke et~al.(2017)Gemmeke, Ellis, Freedman, Jansen, Lawrence, Moore,
  Plakal, and Ritter]{VGGish}
Jort~F Gemmeke, Daniel~PW Ellis, Dylan Freedman, Aren Jansen, Wade Lawrence,
  R~Channing Moore, Manoj Plakal, and Marvin Ritter.
\newblock Audio set: An ontology and human-labeled dataset for audio events.
\newblock In \emph{ICASSP}, pages 776--780. IEEE, 2017.

\bibitem[Georgescu et~al.(2021)Georgescu, Barbalau, Ionescu, Khan, Popescu, and
  Shah]{georgescu2021anomaly}
Mariana-Iuliana Georgescu, Antonio Barbalau, Radu~Tudor Ionescu, Fahad~Shahbaz
  Khan, Marius Popescu, and Mubarak Shah.
\newblock Anomaly detection in video via self-supervised and multi-task
  learning.
\newblock In \emph{CVPR}, pages 12742--12752, 2021.

\bibitem[Ghorbani(2019)]{ghorbani2019mahalanobis}
Hamid Ghorbani.
\newblock Mahalanobis distance and its application for detecting multivariate
  outliers.
\newblock \emph{Facta Universitatis, Series: Mathematics and Informatics},
  pages 583--595, 2019.

\bibitem[Hirschorn and Avidan(2023)]{hirschorn2023normalizing}
Or Hirschorn and Shai Avidan.
\newblock Normalizing flows for human pose anomaly detection.
\newblock In \emph{ICCV}, pages 13545--13554, 2023.

\bibitem[Ioffe and Szegedy(2015)]{BatchNorm}
Sergey Ioffe and Christian Szegedy.
\newblock Batch normalization: Accelerating deep network training by reducing
  internal covariate shift.
\newblock In \emph{ICML}, pages 448--456. pmlr, 2015.

\bibitem[Kingma and Ba(2014)]{kingma2014adam}
Diederik~P Kingma and Jimmy Ba.
\newblock Adam: A method for stochastic optimization.
\newblock \emph{arXiv preprint arXiv:1412.6980}, 2014.

\bibitem[Li et~al.(2022)Li, Liu, and Jiao]{MSL}
Shuo Li, Fang Liu, and LiCheng Jiao.
\newblock Self-training multi-sequence learning with transformer for weakly
  supervised video anomaly detection.
\newblock In \emph{AAAI}, pages 1395--1403, 2022.

\bibitem[Li and Vasconcelos(2015)]{li2015multiple}
Weixin Li and Nuno Vasconcelos.
\newblock Multiple instance learning for soft bags via top instances.
\newblock In \emph{CVPR}, pages 4277--4285, 2015.

\bibitem[Li et~al.(2013)Li, Mahadevan, and Vasconcelos]{li2013anomaly}
Weixin Li, Vijay Mahadevan, and Nuno Vasconcelos.
\newblock Anomaly detection and localization in crowded scenes.
\newblock \emph{PAMI}, 36\penalty0 (1):\penalty0 18--32, 2013.

\bibitem[Liu et~al.(2021{\natexlab{a}})Liu, Nie, Long, Zhang, and
  Li]{liu2021hybrid}
Zhian Liu, Yongwei Nie, Chengjiang Long, Qing Zhang, and Guiqing Li.
\newblock A hybrid video anomaly detection framework via memory-augmented flow
  reconstruction and flow-guided frame prediction.
\newblock In \emph{ICCV}, pages 13588--13597, 2021{\natexlab{a}}.

\bibitem[Liu et~al.(2021{\natexlab{b}})Liu, Ning, Cao, Wei, Zhang, Lin, and
  Hu]{VSwin}
Ze Liu, Jia Ning, Yue Cao, YiXuan Wei, Zheng Zhang, Stephen Lin, and Han Hu.
\newblock Video swin transformer.
\newblock \emph{arXiv preprint arXiv:2106.13230}, 2021{\natexlab{b}}.

\bibitem[Luo et~al.(2017)Luo, Liu, and Gao]{ShanghaiTec}
Weixin Luo, Wen Liu, and Shenghua Gao.
\newblock A revisit of sparse coding based anomaly detection in stacked rnn
  framework.
\newblock In \emph{ICCV}, pages 341--349, 2017.

\bibitem[Markovitz et~al.(2020)Markovitz, Sharir, Friedman, Zelnik-Manor, and
  Avidan]{markovitz2020graph}
Amir Markovitz, Gilad Sharir, Itamar Friedman, Lihi Zelnik-Manor, and Shai
  Avidan.
\newblock Graph embedded pose clustering for anomaly detection.
\newblock In \emph{CVPR}, pages 10539--10547, 2020.

\bibitem[Mehran et~al.(2009)Mehran, Oyama, and Shah]{mehran2009abnormal}
Ramin Mehran, Alexis Oyama, and Mubarak Shah.
\newblock Abnormal crowd behavior detection using social force model.
\newblock In \emph{PAMI}, pages 935--942. IEEE, 2009.

\bibitem[Nguyen and Meunier(2019)]{nguyen2019anomaly}
Trong-Nguyen Nguyen and Jean Meunier.
\newblock Anomaly detection in video sequence with appearance-motion
  correspondence.
\newblock In \emph{ICCV}, pages 1273--1283, 2019.

\bibitem[Pang et~al.(2020)Pang, Yan, Shen, Hengel, and Bai]{pang2020self}
Guansong Pang, Cheng Yan, Chunhua Shen, Anton van~den Hengel, and Xiao Bai.
\newblock Self-trained deep ordinal regression for end-to-end video anomaly
  detection.
\newblock In \emph{CVPR}, pages 12173--12182, 2020.

\bibitem[Park et~al.(2020)Park, Noh, and Ham]{park2020learning}
Hyunjong Park, Jongyoun Noh, and Bumsub Ham.
\newblock Learning memory-guided normality for anomaly detection.
\newblock In \emph{CVPR}, pages 14372--14381, 2020.

\bibitem[Paszke et~al.(2017)Paszke, Gross, Chintala, Chanan, Yang, DeVito, Lin,
  Desmaison, Antiga, and Lerer]{paszke2017automatic}
Adam Paszke, Sam Gross, Soumith Chintala, Gregory Chanan, Edward Yang, Zachary
  DeVito, Zeming Lin, Alban Desmaison, Luca Antiga, and Adam Lerer.
\newblock Automatic differentiation in pytorch.
\newblock 2017.

\bibitem[Qu et~al.(2021)Qu, Mo, and Niu]{qu2021dat}
Yuntao Qu, Shasha Mo, and Jianwei Niu.
\newblock Dat: Training deep networks robust to label-noise by matching the
  feature distributions.
\newblock In \emph{CVPR}, pages 6821--6829, 2021.

\bibitem[Rosenblatt(1956)]{CLT}
Murray Rosenblatt.
\newblock A central limit theorem and a strong mixing condition.
\newblock \emph{Proceedings of the National Academy of Sciences}, 42\penalty0
  (1):\penalty0 43--47, 1956.

\bibitem[Ruff et~al.(2018)Ruff, Vandermeulen, Goernitz, Deecke, Siddiqui,
  Binder, M{\"u}ller, and Kloft]{ruff2018deep}
Lukas Ruff, Robert Vandermeulen, Nico Goernitz, Lucas Deecke, Shoaib~Ahmed
  Siddiqui, Alexander Binder, Emmanuel M{\"u}ller, and Marius Kloft.
\newblock Deep one-class classification.
\newblock In \emph{ICML}, pages 4393--4402. PMLR, 2018.

\bibitem[Sabokrou et~al.(2018)Sabokrou, Khalooei, Fathy, and
  Adeli]{sabokrou2018adversarially}
Mohammad Sabokrou, Mohammad Khalooei, Mahmood Fathy, and Ehsan Adeli.
\newblock Adversarially learned one-class classifier for novelty detection.
\newblock In \emph{CVPR}, pages 3379--3388, 2018.

\bibitem[Srivastava et~al.(2014)Srivastava, Hinton, Krizhevsky, Sutskever, and
  Salakhutdinov]{Dropout}
Nitish Srivastava, Geoffrey Hinton, Alex Krizhevsky, Ilya Sutskever, and Ruslan
  Salakhutdinov.
\newblock Dropout: a simple way to prevent neural networks from overfitting.
\newblock \emph{JMLR}, 15\penalty0 (1):\penalty0 1929--1958, 2014.

\bibitem[Sultani et~al.(2018)Sultani, Chen, and Shah]{UCF}
Waqas Sultani, Chen Chen, and Mubarak Shah.
\newblock Real-world anomaly detection in surveillance videos.
\newblock In \emph{CVPR}, pages 6479--6488, 2018.

\bibitem[Sun et~al.(2021)Sun, Guo, and Li]{sun2021react}
Yiyou Sun, Chuan Guo, and Yixuan Li.
\newblock React: Out-of-distribution detection with rectified activations.
\newblock \emph{NeurIPS}, 34:\penalty0 144--157, 2021.

\bibitem[Tian et~al.(2021)Tian, Pang, Chen, Singh, Verjans, and Carneiro]{RTFM}
Yu Tian, GuangSong Pang, YuanHong Chen, Rajvinder Singh, Johan~W. Verjans, and
  Gustavo Carneiro.
\newblock Weakly-supervised video anomaly detection with robust temporal
  feature magnitude learning.
\newblock In \emph{ICCV}, pages 4955--4966, 2021.

\bibitem[Tran et~al.(2015)Tran, Bourdev, Fergus, Torresani, and Paluri]{C3D}
Du Tran, Lubomir~D. Bourdev, Rob Fergus, Lorenzo Torresani, and Manohar Paluri.
\newblock Learning spatiotemporal features with 3d convolutional networks.
\newblock In \emph{ICCV}, pages 4489--4497, 2015.

\bibitem[Wang et~al.(2018)Wang, Xiong, Wang, Qiao, Lin, Tang, and
  Van~Gool]{TSN}
Limin Wang, Yuanjun Xiong, Zhe Wang, Yu Qiao, Dahua Lin, Xiaoou Tang, and Luc
  Van~Gool.
\newblock Temporal segment networks for action recognition in videos.
\newblock \emph{PAMI}, 41\penalty0 (11):\penalty0 2740--2755, 2018.

\bibitem[Weinberger and Saul(2009)]{TripletLoss}
Kilian~Q Weinberger and Lawrence~K Saul.
\newblock Distance metric learning for large margin nearest neighbor
  classification.
\newblock \emph{JMLR}, 10\penalty0 (2), 2009.

\bibitem[Wilson et~al.(2023)Wilson, Fischer, Dayoub, Miller, and
  S{\"u}nderhauf]{wilson2023safe}
Samuel Wilson, Tobias Fischer, Feras Dayoub, Dimity Miller, and Niko
  S{\"u}nderhauf.
\newblock Safe: Sensitivity-aware features for out-of-distribution object
  detection.
\newblock In \emph{ICCV}, pages 23565--23576, 2023.

\bibitem[Wu et~al.(2022)Wu, Hsieh, Chen, Fuh, and Liu]{S3R}
Jhih-Ciang Wu, He-Yen Hsieh, Ding-Jie Chen, Chiou-Shann Fuh, and Tyng-Luh Liu.
\newblock Self-supervised sparse representation for video anomaly detection.
\newblock In \emph{ECCV}, pages 729--745. Springer, 2022.

\bibitem[Wu et~al.(2020{\natexlab{a}})Wu, Liu, Shi, Shao, Wu, and Yang]{XD}
Peng Wu, Jing Liu, YuJia Shi, FangTao Shao, ZhapYang Wu, and ZhiWei Yang.
\newblock Not only look, but also listen: Learning multimodal violence
  detection under weak supervision.
\newblock In \emph{ECCV}, pages 322--339, 2020{\natexlab{a}}.

\bibitem[Wu et~al.(2020{\natexlab{b}})Wu, Zheng, Goswami, Metaxas, and
  Chen]{wu2020topological}
Pengxiang Wu, Songzhu Zheng, Mayank Goswami, Dimitris Metaxas, and Chao Chen.
\newblock A topological filter for learning with label noise.
\newblock \emph{NeurIPS}, 33:\penalty0 21382--21393, 2020{\natexlab{b}}.

\bibitem[Xie et~al.(2017)Xie, Girshick, Doll{\'a}r, Tu, and He]{ResNeXt}
Saining Xie, Ross Girshick, Piotr Doll{\'a}r, Zhuowen Tu, and Kaiming He.
\newblock Aggregated residual transformations for deep neural networks.
\newblock In \emph{CVPR}, pages 1492--1500, 2017.

\bibitem[Yan et~al.(2023)Yan, Zhang, Liu, Pang, and Wang]{FPDM}
Cheng Yan, Shiyu Zhang, Yang Liu, Guansong Pang, and Wenjun Wang.
\newblock Feature prediction diffusion model for video anomaly detection.
\newblock In \emph{ICCV}, pages 5527--5537, 2023.

\bibitem[Yu et~al.(2022)Yu, Liu, Cheng, Feng, and Zhang]{MACIL-SD}
Jiashuo Yu, Jinyu Liu, Ying Cheng, Rui Feng, and Yuejie Zhang.
\newblock Modality-aware contrastive instance learning with self-distillation
  for weakly-supervised audio-visual violence detection.
\newblock In \emph{ACM MM}, pages 6278--6287, 2022.

\bibitem[Zaheer et~al.(2020)Zaheer, Mahmood, Astrid, and Lee]{CLAWS}
Muhammad~Zaigham Zaheer, Arif Mahmood, Marcella Astrid, and Seung-Ik Lee.
\newblock Claws: Clustering assisted weakly supervised learning with normalcy
  suppression for anomalous event detection.
\newblock In \emph{ECCV}, pages 358--376, 2020.

\bibitem[Zaheer et~al.(2022)Zaheer, Mahmood, Khan, Segu, Yu, and Lee]{GCL}
Muhammad~Zaigham Zaheer, Arif Mahmood, Muhannad~Haris Khan, Mattia Segu, Fisher
  Yu, and Seung-Ik Lee.
\newblock Generative cooperative learning for unsupervised video anomaly
  detection.
\newblock In \emph{CVPR}, pages 14744--14754, 2022.

\bibitem[Zhang et~al.(2023)Zhang, Li, Qi, Wang, Qing, Huang, and Yang]{CUNet}
Chen Zhang, Guorong Li, Yuankai Qi, Shuhui Wang, Laiyun Qing, Qingming Huang,
  and Ming-Hsuan Yang.
\newblock Exploiting completeness and uncertainty of pseudo labels for weakly
  supervised video anomaly detection.
\newblock In \emph{CVPR}, pages 16271--16280, 2023.

\bibitem[Zhong et~al.(2019)Zhong, Li, Kong, Liu, Li, and Li]{GCN}
JiaXing Zhong, NanNan Li, WeiJie Kong, Shan Liu, Thomas~H. Li, and Ge Li.
\newblock Graph convolutional label noise cleaner: Train a plug-and-play action
  classifier for anomaly detection.
\newblock In \emph{CVPR}, pages 1237--1246, 2019.

\bibitem[Zhou et~al.(2023{\natexlab{a}})Zhou, Yu, and Yang]{UR-DMU}
Hang Zhou, Junqing Yu, and Wei Yang.
\newblock Dual memory units with uncertainty regulation for weakly supervised
  video anomaly detection.
\newblock 2023{\natexlab{a}}.

\bibitem[Zhou et~al.(2023{\natexlab{b}})Zhou, Yang, Qu, Xu, Shen, and
  Shen]{AnoOnly}
Yixuan Zhou, Peiyu Yang, Yi Qu, Xing Xu, Fumin Shen, and Heng~Tao Shen.
\newblock Anoonly: Semi-supervised anomaly detection without loss on normal
  data.
\newblock \emph{arXiv preprint arXiv:2305.18798}, 2023{\natexlab{b}}.

\end{thebibliography}
}


\end{document}